%% file: main.tex
\documentclass[10pt, journal]{IEEEtran}
\usepackage{times}
\usepackage{epsfig}
\usepackage{graphicx}
\usepackage{amsmath}
\usepackage{amssymb}
\usepackage{commath}
\usepackage{amsthm}
\usepackage{breqn}
\usepackage{color}
\usepackage{multicol}
\usepackage{arydshln}
\usepackage{url}
\usepackage{algorithmic}
\usepackage[super]{nth}

 \usepackage[norelsize,linesnumbered,longend,ruled,lined,boxed,commentsnumbered]{algorithm2e}
 \usepackage{ragged2e}

\newtheorem{theorem}{Theorem}{}
\newtheorem{corollary}{Corollary}{}
\newtheorem{remark}{Lemma}{}

\usepackage[pagebackref=true,breaklinks=true,colorlinks,bookmarks=false]{hyperref}

\begin{document}

\title{AngularGrad: A New Optimization Technique for Angular Convergence of Neural Networks}

\author{Swalpa Kumar Roy,~\IEEEmembership{Student Member,~IEEE,}
        Mercedes E. Paoletti,~\IEEEmembership{Senior Member,~IEEE,}\\
        Juan M. Haut,~\IEEEmembership{Senior Member,~IEEE,}
        Shiv Ram Dubey,~\IEEEmembership{Senior Member,~IEEE,}
        Purbayan Kar,\\
        Antonio Plaza,~\IEEEmembership{Fellow,~IEEE,}% <-this % stops a space
        ~and
        Bidyut B. Chaudhuri,~\IEEEmembership{Life~Fellow,~IEEE} % <-this % stops a space
\thanks{This publication has been possible thanks to the funding provided by the Consejeria de Economia, Ciencia y Agencia Digital de la Junta de Extremadura and Fondo Europeo de Desarrollo Regional de la Union Europea through reference grant GR21040. This work was also supported in part by the Spanish Ministerio de Ciencia e Innovacion under Project PID2019-110315RB-I00 (APRISA). This research was also supported by Science and Engineering Research Board (SERB), Government of India under Project Grant No. SRG/2022/001390. This work has been supported by the computing facilities of Extremadura Research Centre for Advanced Technologies (CETA-CIEMAT), funded by the European Regional Development Fund (ERDF). CETA-CIEMAT belongs to CIEMAT and the Government of Spain. \textit{(Corresponding author: Juan M. Haut)}}
\thanks{S.K. Roy is with the Computer Science and Engineering at Alipurduar Government Engineering and Management College, 736206, India.}% <-this % stops an unwanted space
\thanks{M.E. Paoletti, J.M. Haut and A. Plaza is with the Hyperspectral Computing Laboratory, Department of Technology of Computers and Communications, University of Extremadura, 10003 Cáceres, Spain.}% <-this % stops an unwanted space
\thanks{S.R. Dubey is with the Computer Vision and Biometrics Lab, Indian Institute of Information Technology, Allahabad, Prayagraj, Uttar Pradesh-211015, India.}% <-this % stops an unwanted space
\thanks{P. Kar is with the Media Analysis Group, Sony Research India Private Limited, Bangalore, Karnataka 560103, India.}
\thanks{B.B. Chaudhuri is with the Computer Vision and Pattern Recognition Unit at Indian Statistical Institute, Kolkata-700108, India.}% <-this % stops an unwanted space
}

\maketitle

% \markboth{Submitted to IEEE journal}%
% {Roy \MakeLowercase{\textit{et al.}}: AngularGrad: A New Optimization Technique for Angular Convergence of Convolutional Neural Networks}

\begin{abstract}
Convolutional neural networks (CNNs) are trained using stochastic gradient descent (SGD)-based optimizers. Recently, the adaptive moment estimation (Adam) optimizer has become very popular due to its adaptive momentum, which tackles the dying gradient problem of SGD. Nevertheless, existing optimizers are still unable to exploit the optimization curvature information efficiently. This paper proposes a new AngularGrad optimizer that considers the behavior of the direction/angle of consecutive gradients. This is the first attempt in the literature to exploit the gradient angular information apart from its magnitude. The proposed AngularGrad generates a score to control the step size based on the gradient angular information of previous iterations. Thus, the optimization steps become smoother as a more accurate step size of immediate past gradients is captured through the angular information. Two variants of AngularGrad are developed based on the use of Tangent or Cosine functions for computing the gradient angular information. Theoretically, AngularGrad exhibits the same regret bound as Adam for convergence purposes. Nevertheless, extensive experiments conducted on benchmark data sets against state-of-the-art methods reveal a superior performance of AngularGrad. Source code: \url{https://github.com/mhaut/AngularGrad}.
\end{abstract}

\begin{IEEEkeywords}
Adam, Gradient Descent, Image Classification, Neural Networks, Optimization.
\end{IEEEkeywords}

\section{Introduction}\label{sec:introduction}
\IEEEPARstart{D}{uring} the last decades, remarkable technological advances have provided unprecedented computing and storage enhancements by reducing transistor geometries whilst increasing the complexity of integrated circuits \cite{moore1998cramming}. On the one hand, this has yielded outstanding improvements in data collection devices. Indeed, both hardware and software elements implemented into current sensor systems offer unparalleled data acquisition ratios. As a result, this has entailed a vast increase in the size of the collected (big) data sets, which now contain increasingly complex and rich information for diverse applications \cite{chen2014data}. In this context, there is a compelling requirement to process such data fast enough for practical use, in order to extract relevant information to improve many tasks which rely on data-driven decision processes.

% https://www.kdnuggets.com/2017/05/deep-learning-big-deal.html
Computer advances have indeed facilitated the development of new and powerful processing tools. Since 2016, remarkable improvements in the processing power provided by graphical processing units (GPUs) has led to the rise of deep learning (DL) models \cite{lecun2015deep,goodfellow2016deep}. DL models evolved from standard artificial neural networks (ANNs) in the field of machine learning (ML) \cite{bishop2006pattern}, to a current scenario in which ANNs mimic the human brain by adapting to new events and stimuli, and are based on nodes organised and interconnected through synaptic weights in a hierarchical layered structure. Each neuron in a particular layer receives the input data, which are consequently processed through the weights of the neurons. As a result, by properly adjusting the involved parameters (and without any prior information about data distribution), the data are processed and refined along the network, resulting in an abstract representation that entails the most relevant information for the final interpretation task \cite{gopisetty2008evolution}. Therefore, with the availability of powerful computing devices based on multi-core CPUs and GPUs, the development of increasingly complex and deep neural models is now possible. \cite{schmidhuber2015deep}. 

In fact, DL architectures can also be interpreted as a deep stack of different layers, which are composed of computing units such as standard neurons, recurrent cells, spatial kernels or capsules \cite{sabour2017dynamic}. These units conduct different transformations to the input layers by applying a series of operations to the data, according to some parameters (weights and biases) and depending on the layer to which they belong. Layers can be connected in a standard, fully connected (FC) way (for instance, multilayer perceptrons --MLPs-- \cite{cirecsan2012deep} or FC-based models such as stacked autoencoders --SAEs-- \cite{zabalza2016novel} or deep belief networks --DBNs-- \cite{hinton2009deep}), or by local connections (such as convolutional neural networks --CNNs-- \cite{NIPS2012_c399862d}) and recurrent links (as those exhibited by recurrent networks --RNNs-- \cite{pascanu2013difficulty}). Furthermore, layers of different levels can be connected through skip and residual links (residual --ResNet-- \cite{he2016deep} and dense networks --DenseNet-- \cite{huang2017densely}). In this context, DL models offer great flexibility in terms of architecture design, providing different degrees of structural complexity, where deeper and more complex models can combine different data paths \cite{szegedy2016rethinking,wang2017residual,chen2017dual} and even models \cite{goodfellow2014generative,zhang2019deeper,mirzaei2020deep}. As a result, a wide range of DL models have obtained outstanding results in a wide range of hot-topic research challenges, such as computer vision \cite{NIPS2012_c399862d}, signal processing \cite{yu2010deep}, natural language identification \cite{young2018recent}, object detection \cite{szegedy2013deep} and tracking \cite{wang2013learning}, robotics \cite{pierson2017deep}, image segmentation \cite{wang2018interactive} and classification \cite{paoletti2019deep}, among others. 

However, training these models entails a number of challenges in addition to the associated computational burden, that increases as the architecture deepens. Indeed, deep networks require a large amount of training data (which must also satisfy variability) to perform accurately, and they are prone to overfitting. In this sense, and despite the large amount of data available, the number of training instances is often limited, as the labelling and preparation of training samples is quite expensive and time consuming in many application domains. Moreover, as the architecture becomes deeper, the gradients tend to vanish. This hinders the learning process \cite{he2016deep}. In the following, we delve into these drawbacks, particularly focusing on the optimization of CNNs \cite{NIPS2012_c399862d,simonyan2014very,he2016deep,xie2017aggregated,huang2017densely}.

\subsection{Motivation}
Mathematically, a DNN can be interpreted as a mapping function $f(\cdot,\theta)$ which matches an input $x\in\mathcal{X}$ with the corresponding output $y\in\mathcal{Y}$, i.e., $f:\mathcal{X}\rightarrow \mathcal{Y}$, by adjusting its parameters $\theta$ (weights and biases) to optimize a cost or loss function $J(\theta)$. The broad success of this model relies on the fact that $f$ works as an universal approximation function when given an appropriate network configuration in terms of depth and weights \cite{csaji2001approximation,zhou2020universality}. Although the depth of the model depends on several aspects, the weights can be optimally achieved through an iterative process by means of which the model automatically adjusts the values of its weights through a forward-backward procedure to reach the minimum value of $J(\theta)$. Thus, the optimizer plays a key role in the efficiency of the training process and the final generalization performance of the DNN \cite{yong2020gradient}. 

%\subsection{Problem Statement}
Many efforts have been devoted to developing optimization algorithms with a good trade-off between training acceleration and model generalization. In particular, the simple and effective gradient descent method has been widely explored, providing a variety of algorithms \cite{bottou1991stochastic,qian1999momentum,hinton2012neural,duchi2011adaptive,kingma2014adam,loshchilov2017decoupled,dubey2019diffgrad,yong2020gradient}. Among these, stochastic gradient descent (SGD) \cite{bottou1991stochastic} and its momentum version (SGDM) \cite{qian1999momentum} are quite popular due to their simplicity, although these methods (SGD in particular) are highly affected by the high variance in model parameters and the vanishing gradient problem. To overcome these limitations, some algorithms attempt to produce an adaptive learning rate, such as RMSProp \cite{hinton2012neural} and Adam \cite{kingma2014adam}. Indeed, most of the algorithms differ in the degree of information they take into account when calculating the next step, such as diffGrad \cite{dubey2019diffgrad} and gradient centralization (GC) \cite{yong2020gradient}.  In addition, some approaches combine optimization algorithms with other external techniques to improve the learning process, such as the dynamic update of the learning rate \cite{baydin2017online}, the appropriate setting of the model hyperparameters \cite{li2020rethinking}, or the search for an optimal region for model initialization \cite{dauphin2019metainit}. However, these models cannot effectively tackle the zig-zag phenomenon (i.e., noisy updates) in the optimization trajectory caused by a high variation of the gradients. As a result, the convergence curve usually exhibits abrupt fluctuations that hinder the final performance.

\subsection{Contributions}
To overcome the aforementioned limitations, this paper introduces \texttt{AngularGrad} as a new and effective optimization algorithm, which reduces the zig-zag effect in the optimization trajectory and speeds up convergence. 
% \REMOVE{by taking into account the angle between two consecutive gradients during optimization steps}. 
To implement our new optimizer, we utilize the direction/angle information of the gradient vector. To the best of our knowledge, only the magnitude of the gradient has been used for optimization in the literature. This is the first time that both direction and angle information are jointly considered to improve the optimization process. Here, the angle between two consecutive gradient direction iterations is utilized. As a result, trajectory fluctuations are significantly smoothed, tracing a more direct path towards the minimum of the cost function. Our method also reduces the need for computational resources. The proposed \texttt{AngularGrad} has been extensively evaluated on challenging data sets using a wide variety of CNN architectures. We have also conducted comparisons with state-of-the-art optimization methods to prove its effectiveness and efficiency. 

The remainder of the paper is organized as follows. Section \ref{relatedworks} reviews popular optimization algorithms, analyzing their limitations. Section \ref{proposed} provides methodological details about our newly proposed \texttt{AngularGrad}. Sections \ref{sec:convergence} and \ref{sec:empirical} discuss the convergence and empirical properties of \texttt{AngularGrad}. Section \ref{sec:experiments} provides an extensive evaluation, analyzing the performance of our method in comparison with that of other state-of-the-art optimizers via several experiments. Section \ref{concl} concludes the paper with some remarks and hints at plausible future research lines.

%%%%%%%%%%%%%%%%%%%%%%%%%%% %%%%%%%%%%%%%%%%%%%%%%%%%%% %%%%%%%%%%%%%%%%%%%%%%%%%%%
%%%%%%%%%%%%%%%%%%%%%%%%%%% %%%%%%%%%%%%%%%%%%%%%%%%%%% %%%%%%%%%%%%%%%%%%%%%%%%%%%
%%%%%%%%%%%%%%%%%%%%%%%%%%% %%%%%%%%%%%%%%%%%%%%%%%%%%% %%%%%%%%%%%%%%%%%%%%%%%%%%%
\section{Related Works}
\label{relatedworks}
The SGD is a widely used optimizer that faces several challenges. These can be summarized as follows: i) choosing a proper learning rate is quite hard; ii) the same learning rate in an epoch is used for updating all parameters; and iii) it tends to get stuck at local minima during the optimization process. To address iii), a momentum factor has been introduced which accelerates the SGD in the relevant direction and diminishes oscillations (resulting in the SGDM) \cite{sutskever2013importance, smith2021origin}. It adds a fraction $\gamma$ of the update vector from the past time step to the current update vector. The parameter update can be written following Eq. (\ref{eq:sgdparamupdate}):
\begin{equation}
    \theta_{t} = \theta_{t-1} - \alpha_{t}m_t,
    \label{eq:sgdparamupdate}
\end{equation}
where $\theta_{t-1}$ and $\theta_{t}$ are the original and updated parameters after the $t^{th}$ iteration, respectively, $\alpha_t$ is the step size or learning rate, and $m_t$ is the gradient based on parameter $\theta$ and computed as:
\begin{equation}
    m_{t} = \gamma m_{t-1}+ (1-\gamma) \bigtriangledown_{\theta}\mathcal{L}(\theta_{t}),
\end{equation}
where $\mathcal{L}$ is the loss function defined with respect to the parameters of the model, $m_{t}$ is the momentum at the $t^{th}$ iteration with $m_0=0$, $\gamma$ is a constant, and $\mathbf{g_t} = \bigtriangledown_{\theta}\mathcal{L}(\theta_{t})$ defines the gradient for parameter $\theta$. As it can be observed, the gradient is included to gain moment for the parameters having consistent gradient \cite{sutskever2013importance}.

However, an adaptive learning rate is necessary to tackle those problems related to a constant learning rate. In this regard, the Adam optimizer has been specifically developed to deals with them~\cite{kingma2014adam}. It stores both an exponentially decaying average of past gradients $m_t$ and an exponentially decaying average of the past squared gradients $v_t$. Indeed, $m_t$ and $v_t$ are defined as the $1^{st}$ and $2^{nd}$ order moments:
\begin{equation}
    m_{t} = \beta_{1}m_{t-1} + (1-\beta_{1})\mathbf{g_{t}},
\end{equation}
\begin{equation}
    v_{t} = \beta_{2}v_{t-1} + (1-\beta_{2})\mathbf{g_{t}^2},
\end{equation}
where $\beta_{1}$ and $\beta_{2}$ are hyperparameters to control the exponential decay rates, $m_{t-1}$ is the mean of past gradients and $v_{t-1}$ the variance. A bias correction is done by the Adam optimizer to get rid of a very large step size in the initial iterations: $\hat{m}_t = \frac{m_{t}}{1-\beta_1^{t}}$ and $\hat{v_{t}} = \frac{v_{t}}{1-\beta_2^{t}}$.
% \begin{equation}
%     \hat{m}_t = \frac{m_{t}}{1-\beta_1^{t}}~{\rm and}~\hat{v_{t}} = \frac{v_{t}}{1-\beta_2^{t}}
% \end{equation}
% \begin{equation}
%     \hat{v_{t}} = \frac{v_{t}}{1-\beta_2^{t}}
% \end{equation}
The parameter update in Adam is performed as follows:
\begin{equation}
    \theta_{t} = \theta_{t-1} - \frac{\alpha_{t}}{\sqrt{\hat{v_t}}+\epsilon}\hat{m_t},
    \label{eq:AdamparamUpdate}
\end{equation}
where the values of $\beta_1$, $\beta_2$ are usually set to $0.9$ and $0.999$, whilst parameter $\epsilon$ is introduced to avoid the case of division by zero with a value normally set to $10^{-8}$. As in Eq. (\ref{eq:sgdparamupdate}), $\alpha_t$ is the learning rate. In this regard, Adam with decoupled weight decay (AdamW) \cite{loshchilov2017decoupled} extends Adam, where the update rule is given by Eq. (\ref{eq:AdamWparamUpdate}):
\begin{equation}
    \theta_{t} = \theta_{t-1} - \frac{\alpha_{t}}{\sqrt{\hat{v_t}}+\epsilon}\hat{m_t}+\lambda \theta_{t-1},
    \label{eq:AdamWparamUpdate}
\end{equation}
where $\lambda$ is a constant.

\begin{algorithm}[!t]
\caption{diffGrad Optimizer}
\SetKwInOut{Initialize}{initialize}
\Initialize{$\theta_{0},m_{0}\gets0,v_{0}\gets0,t\gets0$}
\begin{algorithmic}
\WHILE{$\theta_{t}$ not converged}
    \STATE $t \gets t+1$
    \STATE $\mathbf{g_t} \gets \nabla_{\theta} f_t(\theta_{t-1})$
    \STATE $\Delta \mathbf{g_{t}} \gets \mathbf{g_{t}} - \mathbf{g_{t-1}}$
    \STATE {$\xi_{t} \gets AbsSig(\Delta \mathbf{g_{t}})$}
    \STATE $m_t \gets \beta_1 \cdot m_{t-1} + (1-\beta_1) \cdot \mathbf{g_t}$
    \STATE $v_t \gets \beta_2 \cdot v_{t-1} + (1-\beta_2) \cdot \mathbf{g^2_t}$
    \STATE $\widehat{m_t} \gets m_t / (1-\beta_1^t)$ 
    \STATE $\widehat{v_t} \gets v_t / (1-\beta_2^t)$
    \STATE $\theta_t \gets \theta_{t-1} - \alpha \cdot \xi_{t} \cdot \widehat{m}_t$/ $(\sqrt{\widehat{v_t}} + \epsilon)$
\ENDWHILE
\RETURN $\theta_t$
\end{algorithmic}
\label{alg:diffgrad}
\end{algorithm}

Adaptive learning rate-based algorithms face the risk of converging to local optima, due to the high variance of data during training. Such variance can be reduced by a warm-up before optimization process. But the degree of warm-up is generally unknown and varies from one data set to another. Thus, RAdam \cite{liu2019radam} introduces a variance reduction term to automatically control the variance during training, where
$\rho_{\infty} = 2/(1-\beta_{2})-1$ and $\rho_{t} = \rho_{\infty}-2t\beta^{t}_{2}/(1-\beta^{t}_{2})$ are calculated, and the variance reduction term is defined by Eq. (\ref{eq:RAdamvariance}):
\begin{equation}
    r_{t} = \sqrt{\frac{(\rho_{t}-4)(\rho_{t}-2)\rho_{\infty}}{(\rho_{\infty}-4)(\rho_{\infty}-2)\rho_{t}}}
    \label{eq:RAdamvariance}
\end{equation}
In particular, when $\rho_{t}>4$, the update is given as follows:
\begin{equation}
    \theta_{t}=\theta_{t-1}-\alpha_{t}r_{t}\hat{m}_{t}/\hat{v}_{t}
\end{equation}
Otherwise, the parameter update is conducted exactly as SGD does. 

Notwithstanding the mechanisms introduced to avoid a constant learning rate, one shortcoming of Adam optimizer is that it does not take into account the phenomena of changing gradients in the optimization steps~\cite{kingma2014adam}. To overcome this limitation, the diffGrad~\cite{dubey2019diffgrad} optimizer updates its parameters in such a way that it exhibits a larger step size for those gradient parameters that change rapidly, and a lower step size for those gradient parameters that do not significantly change. With this purpose, the authors introduced a so-called diffGrad friction coefficient $\xi$, which is obtained by analyzing the change of gradients between the present and past iterations as follows:
\begin{equation}
    \xi_{t} = \frac{1}{1+e^{-|\mathbf{g_{t-1}} - \mathbf{g_{t}}|}}
\end{equation}
% \begin{equation}
%     \xi_{t} = \frac{1}{1+e^{-|g_{t,i} = g_{t-1,i} - g_{t,i}|}} AbsSig(\Delta g_{t,i})
% \end{equation}
% \begin{equation}
%     AbsSig(x) = \frac{1}{1+e^{-\abs{x}}} 
% \end{equation}
% where $AbsSig(x) = \frac{1}{1+e^{-\abs{x}}}$, and $\Delta g_{t,i}$ is the change of gradients between present and past iterations given as $\Delta g_{t,i} = g_{t-1,i} - g_{t,i}$
% \begin{equation}
%     \Delta g_{t,i} = g_{t-1,i} - g_{t,i}
% \end{equation}
Then, the diffGrad parameters are updated by Eq. (\ref{eq:diffGradparamUpdate}):
\begin{equation}
    \theta_{t} = \theta_{t-1} - \frac{\alpha_t\xi_{t}}{\sqrt{\hat{v_t}}+\epsilon}\hat{m_t}
    % carefull, alpha_t doesn't exist in algorithm 1
    \label{eq:diffGradparamUpdate}
\end{equation}
Here, $t$ stands for the $t^{th}$ iteration. A detailed description of diffGrad is given in Algorithm~\ref{alg:diffgrad}. 
% and $i$ stands for $i^{th}$ parameter. 
%The algorithm of diffGrad optimizer is shown in Algorithm \ref{alg:diffgrad}. 
% \begin{algorithm}
% \caption{diffGrad Optimizer}
% \SetKwInOut{Initialize}{initialize}
% \Initialize{$\theta_{0},m_{0}\gets0,v_{0}\gets0,t\gets0$}
% \begin{algorithmic}
% \WHILE{$\theta_{t}$ not converged}
%     \STATE $t \gets t+1$
%     \STATE $g_t \gets \nabla_{\theta} f_t(\theta_{t-1})$
%     \STATE $\Delta g_{t} \gets g_{t} - g_{t-1}$
%     \STATE \textcolor{blue}{$\xi_{t} \gets AbsSig(\Delta g_{t})$}
%     \STATE $m_t \gets \beta_1 \cdot m_{t-1} + (1-\beta_1) \cdot g_t$
%     \STATE $v_t \gets \beta_2 \cdot v_{t-1} + (1-\beta_2) \cdot g^2_t$
%     \STATE $\widehat{m_t} \gets m_t / (1-\beta_1^t)$ 
%     \STATE $\widehat{v_t} \gets v_t / (1-\beta_2^t)$
%     \STATE $\theta_t \gets \theta_{t-1} - \alpha \cdot \xi_{t} \cdot \widehat{m_t} \cdot$/ $(\sqrt{\widehat{v_t}} + \epsilon)$
% \ENDWHILE
% \RETURN $\theta_t$
% \end{algorithmic}
% \label{alg:diffgrad}
% \end{algorithm}
% -------------------------------------------

\begin{algorithm}[!t]
\caption{\texttt{AngularGrad} Optimizer}
\SetKwInOut{Initialize}{initialize}
\Initialize{$\theta_{0},m_{0}\gets0,v_{0}\gets0,t\gets0,\lambda_{1}=\frac{1}{2} ,\lambda_{2}=\frac{1}{2}, \sphericalangle=\{\measuredangle\cos, \measuredangle\tan\}$}
\begin{algorithmic}
\WHILE{$\theta_{t}$ not converged}
    \STATE $t \gets t+1$
    \STATE $\mathbf{g_t} \gets \nabla_{\theta} f_t(\theta_{t-1})$
    % \STATE $\measuredangle\tan A_t \gets \abs{(g_t - g_{t-1}) / (1 + g_t\cdot g_{t-1})}$
    % \STATE $\measuredangle\cos A_t \gets 1/\sqrt{1 + \measuredangle\tan^2 A_t}$
    \STATE \textcolor{blue}{$A_t \gets {\tan}^{-1}\abs{(\mathbf{g_t} - \mathbf{g_{t-1}}) / (1 + \mathbf{g_t}\cdot \mathbf{g_{t-1}})}$}
    \STATE \textcolor{blue}{$A_{min} \gets \min(A_{t-1},A_{t})$}
    \STATE \textcolor{blue}{${\phi_t} \gets \tanh{(|\sphericalangle(A_{min})|)}\cdot \lambda_{1} + \lambda_{2}$} 
    \STATE $m_t \gets \beta_1 \cdot m_{t-1} + (1-\beta_1) \cdot \mathbf{g_t}$
    \STATE $v_t \gets \beta_2 \cdot v_{t-1} + (1-\beta_2) \cdot \mathbf{g^2_t}$
    \STATE $\widehat{m_t} \gets m_t / (1-\beta_1^t)$ 
    \STATE $\widehat{v_t} \gets v_t / (1-\beta_2^t)$
    \STATE $\theta_t \gets \theta_{t-1} - \alpha \cdot {\phi_t} \cdot \widehat{m}_t \cdot$/ $(\sqrt{\widehat{v_t}} + \epsilon)$
\ENDWHILE
\RETURN $\theta_t$
\end{algorithmic}
\label{alg:prop}
\end{algorithm}

\begin{figure*}[!htbp]
    \centering
    \includegraphics[width=0.9\linewidth]{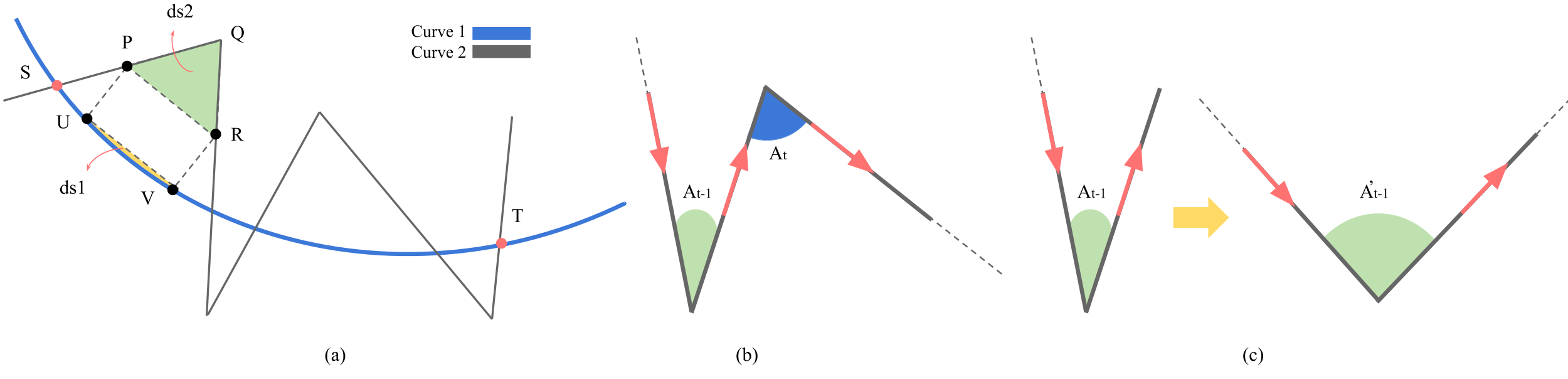}
    \caption{%(a) if an optimizer follows Curve 1 rather than Curve 2 it will have to cover larger distance to reach to global minima. Hence the net number of epochs will be greater. (b) between two consecutive angles $\theta_{1}$ and $\theta_{2}$, \texttt{AngularGrad} takes $\theta_{1}$ (the smaller angle) to calculate ${\texttt{ angular\_co-efficient}}(\phi)$ (c) which makes the angle($\theta_{1}$) flat and hence makes the convergence curve smooth. 
    Graphical illustration of angular gradient information. 
    %\textcolor{red}{(Add little more information here for (a), (b) and (c)).......}
    {The proposed optimizer smoothes the gradient path to accelerate its convergence by flattening the angle between consecutive gradients. Specifically, our \texttt{AngularGrad} optimizer flattens the steep angle (say A) between two vectors and converts it into A' which results in a smoothing of the curve.}
    }
    \label{fig:ang_proof}
\end{figure*}

GC \cite{yong2020gradient} was introduced to remove the mean from the gradient vectors, centralizing those to have a zero mean. It can be used with any optimizer for better convergence. The GC operator is given as follows:
\begin{equation}
    \Phi_{GC}(\bigtriangledown w_{i}\mathcal{L}) = \bigtriangledown w_{i}\mathcal{L} - \mu \bigtriangledown w_{i}\mathcal{L},
\end{equation}
{where $\mathcal{L}$ is the loss function, $\bigtriangledown w_{i}$ %\mathcal{L}$ 
is the gradient of the $i^{th}$ weight, $\mu \bigtriangledown w_{i}\mathcal{L}=\frac{1}{M}\sum_{j=1}^{M}\bigtriangledown w_{i,j}\mathcal{L}$ --$M$ is the dimension-- and $\Phi_{GC}(\bigtriangledown w_{i}\mathcal{L})$ is the centralized gradient.} On the other hand, instead of using the exponential moving average (EMA) of $\mathbf{g^{2}_t}$, AdaBelief \cite{zhuang2020adabelief} uses the EMA of $(\mathbf{g_t}-m_t)^2$ as $s_t$ and the update direction for AdaBelief is $m_t/\sqrt{s_t}$.

The hypergradient descent (HGD) \cite{baydin2017online} tries to derive an update rule for the learning rate $\alpha$ itself. Making an assumption that optimal value of $\alpha$ does not change much between two consecutive iterations, it uses the update rule for the previous step to optimize $\alpha$ in the current one. The HGD is given by:
\begin{equation}
\label{eq:}    \theta_{t} = \theta_{t-1} - \alpha_t \bigtriangledown \mathcal{L}(\theta_{t-1}),
\end{equation}
where $\alpha_{t}=\alpha_{t-1}-\omega\frac{\delta \mathcal{L}(\theta_{t-1})}{\delta \alpha} = \alpha_{t-1}+\omega \bigtriangledown \mathcal{L}(\theta_{t-1}) \cdot \bigtriangledown \mathcal{L}(\theta_{t-2})$ and $\omega$ is the hypergradient learning rate.

At this point, it is important to note that there are quite a few other optimizers\footnote{https://pypi.org/project/torch-optimizer/}. However, a large number of them (e.g. SGD, SGDW, AdaBelief, Lookahead, etc.) cannot reach the global optima in a stipulated number of iterations as per the multi-optima and single optimum benchmark functions, namely Rosenbrock and Rastrigin\footnote{https://github.com/jettify/pytorch-optimizer}.

\section{Proposed \texttt{AngularGrad} Optimization}
\label{proposed}
This section presents our proposed optimization method, which is inspired by diffGrad. Although this method exploits the change of gradients during optimization steps, it is unable to completely remove the high variance of the gradients during intermediate steps. Thus, the convergence curve (though better than Adam) still exhibits a zigzagging pattern. To solve this issue, we propose a new \texttt{AngularGrad} optimizer, which takes into account the angle between two consecutive gradients during optimization. Specifically, we introduce a new angular coefficient ($\phi_t$) which is calculated as follows:
\begin{equation}
    {\phi_t} = \tanh{(|\sphericalangle (A_{min})|)\cdot \lambda_{1}} + \lambda_{2},
    \label{equ:angcoeff}
\end{equation}
where $\{\lambda_{1}, \lambda_{2}\} \in \{0,1\}$ are hyperparameters. We have found empirically that the best value for both $\lambda_{1}$ and $\lambda_{2}$ is $\frac{1}{2}$. Here, $\sphericalangle$ represents either $\measuredangle\cos$ or $\measuredangle\tan$, while $tanh(x)$ is a non-linear function that squashes all values of $x$ between $-1$ and $1$ as follows:
\begin{equation}
    tanh(x) = \frac{\exp^{x}-\exp^{-x}}{\exp^{x}+\exp^{-x}},
\end{equation}
and $A_{t}$ is the angle between gradients at consecutive steps $g_{t}$ and $g_{t-1}$ [illustrated in Fig.~\ref{fig:ang_proof}(b)]. Similarly, the angle between gradients at the $(t-1)^{th}$ step is calculated and termed as $A_{t-1}$. So, $A_{min}$ is computed as follows:
\begin{equation}
    A_{min} = \min(A_{t-1},A_{t}).
\end{equation}
The angular coefficient $\phi_t$ is used to dynamically adjust the learning rate. Our new \texttt{AngularGrad} not only ensures that parameter updates should be smaller in low-gradient changing regions and vice-versa, but also reduces the high variance of the gradients as it minimizes the direction cosines of two consecutive gradients in each step. To do so, our \texttt{AngularGrad} computes first and second order moments, namely $m_{t}$ and $v_{t}$, and bias-corrected moments $\hat{m}_{t}$ and $\hat{v_{t}}$, similar to Adam and diffGrad. Thus, the parameter update using the proposed AngularGrad is done as follows:
\begin{equation}
    \theta_{t} = \theta_{t-1} - \frac{\alpha_{t}\cdot {\phi_t}}{\sqrt{\hat{v_t}}+\epsilon}\widehat{m_t}.
\end{equation}

Algorithm \ref{alg:prop} provides a detailed description of \texttt{AngularGrad}. Depending on the values of coefficient $\phi_t$ in Eq.~(\ref{equ:angcoeff}) --which is evaluated based on the angles $\measuredangle\cos$ and $\measuredangle\tan$, respectively-- we propose two versions of the optimizer, denoted as \texttt{AngularGrad$^{\cos}$} and \texttt{AngularGrad$^{\tan}$}, respectively. To illustrate the improved convergence of \texttt{AngularGrad} optimizer, the following Lemmas are provided:

\begin{remark}
Less time is required for convergence if the path of the curve is smoother rather than zigzagging. %noisy (zig-zag).
\end{remark}
\begin{proof}
Let us consider the curves in Figs.~\ref{fig:ang_proof}(a), i.e., Curve 1 and Curve 2 where Curve 1 is smoother than the Curve 2. Also, let us take 2 points ($S$ and $T$) between the intersection of the two curves. We can draw a triangle $\Delta PQR$ between the area of intersection of the two curves, where point $PQR$ are lies on the Curve 2. Now, according to the triangle inequality statement, we observe that $PQ+QR>PR \implies$ area covered by $\Delta PQR >$ area covered by $PR$. Thus, it can also be shown that the area covered by $\Delta PQR >$ area covered by $UV$. It is also evident that the perimeter covered by $\Delta PQR >$ the perimeter covered by $UV$. Hence, it can be proved that, if an optimizer follows Curve 1 rather than Curve 2, its convergence time will be reduced significantly. 
\end{proof}

\begin{remark}
Fewer epochs are needed to reach the minimum if the path of the curve is smoother rather than zigzagging.%noisy (zig-zag). 
\end{remark}
\begin{proof}
Let us denote $ds_{1}$ and $ds_{2}$ as infinitesimally small distances on Curve 1 and infinitesimally small triangle $\Delta PQR$ on Curve 2 in Fig.~\ref{fig:ang_proof}, respectively. From Fig.~\ref{fig:ang_proof}(a) and triangle inequality, we can see that $PQ + QR > PR$. Now clearly, $PQ + QR >$ perimeter of UV. Let, $dp = PQ + QR$ and $dp = ds_{1}$. Now consider the same speed $dv$ for both Curves 1 and 2.  
%$ds_{2}>ds_{1}$, as $PQ=ds_{2}$, is the hypotenuse of the right-angled triangle $\Delta PQR$. Also, consider the same speed $dv$ for both Curves 1 and 2. 
Hence, the number of steps to cover both distances can be computed as $\frac{ds_{1}}{dv} = dt_{1}$ and $\frac{dp}{dv} = dt_{2}$, respectively. Then, we can write the following relation:
\begin{equation}
    \begin{split}
        dp>ds_{1} & \implies \frac{dp}{dv}>\frac{ds_{1}}{dv} \implies dt_{2}>dt_{1}.
        \label{equ:proof_21}
    \end{split}
\end{equation}
This relation holds between the total distance, from $S$ to $T$:
\begin{equation}
    \int_{S}^{T}dp>\int_{S}^{T}ds_{1}.
    \label{equ:proof_22}
\end{equation}
Therefore, the path traversed by Curve 2 is larger than that traversed by Curve 1. We can write the following with the help of  Eqs.~(\ref{equ:proof_21}) and (\ref{equ:proof_22}):
\begin{equation}
    \begin{aligned}
        dt_{2}>dt_{1} & \implies \int_{S}^{T}dt_{2}>\int_{S}^{T}dt_{1} \implies t_{2}>t_{1}.
    \end{aligned}
\end{equation}
Thus, fewer epochs are required to reach the minimum, and hence the function followed by Curve 1 will take less time to reach the final point $T$ from point $S$.
\end{proof}

% \begin{figure*}[!ht]
% \centering
% \resizebox{0.95\linewidth}{!}{
% \begin{tabular}{ccc}
% \includegraphics[width=\linewidth]{figures/function_0.png} &
% \includegraphics[width=\linewidth]{figures/regression_0.png} &
% \includegraphics[width=\linewidth]{figures/deriv_0.png} \\
% \huge (a) & \huge (b) & \huge (c) \\
% [8pt]
% \includegraphics[width=\linewidth]{figures/function_1.png} &
% \includegraphics[width=\linewidth]{figures/regression_1.png} &
% \includegraphics[width=\linewidth]{figures/deriv_1.png} \\
% \huge (d) & \huge (e) & \huge (f) \\
% [8pt]
% \includegraphics[width=\linewidth]{figures/function_2.png} &
% \includegraphics[width=\linewidth]{figures/regression_2.png} &
% \includegraphics[width=\linewidth]{figures/deriv_2.png} \\
% \huge (g) & \huge (h) & \huge (i) \\
% \end{tabular}}
% \caption{Empirical results over toy examples using SGDM, Adam, diffGrad, AdaBelief, \texttt{AngularGrad$^{\cos}$} and \texttt{AngularGrad$^{\tan}$}.}
% \label{fig:analytical}
% \end{figure*}

\begin{remark}
Angular coefficient makes the updates invariant to sharp curvature changes.
\end{remark}
\begin{proof}
As $\cos{0}=1$, $\cos{\frac{\pi}{2}}=0$ and $0 \leq \abs{\cos{A}} \leq 1$, then $0.5 \leq \frac{\tanh{\abs{\cos{A}}}}{2}+\frac{1}{2} \leq 1$, where $A$ is the angle between two consecutive gradients. So, $0.5 \leq \phi \leq 1$. We can write the following statements:
\begin{itemize}
    \item if $A \approx 0 \implies$ angular coefficient, $\phi \approx 1$.
    \item if $A \approx \frac{\pi}{2} \implies$ angular coefficient, $\phi \approx 0.5$. 
\end{itemize}
% \noindent When $\theta \approx 0$,
% \begin{equation}
%     \begin{split}
%         {{angular\_coefficient}}, \phi \approx 1
%     \end{split}
% \end{equation}
% When $\theta \approx \frac{\pi}{2}$, 
% \begin{equation}
%     \begin{split}
%         {{angular\_coefficient}}, \phi \approx 0.5
%     \end{split}
% \end{equation}
% And it applies for all the multiples of $\theta$. 
In this regard, when the angle is steep, $\phi$ must be higher $\implies$ learning rate must be higher to flatten the angle $\implies$ the path becomes smoother (and vice-versa). The proposed \texttt{AngularGrad} will compute $\min(A_{t-1},A_{t}) = A_{t-1}$ as $A_{t-1} \approx 0$, following Fig.~\ref{fig:ang_proof}(b), which leads to:
\begin{equation}
    \begin{split}
        \mbox{angular coefficient}, \phi \approx 1 \textbf{ } {\rm (Larger~value)}.
    \end{split}
\end{equation}
Hence, the curve becomes flat, as shown in Fig.~\ref{fig:ang_proof}(c). Our proposed \texttt{AngularGrad} optimizer flattens the steep angle (say A) between two vectors, and converts it into A', which results in a smoothing of the curve.
\end{proof}

\section{Convergence Analysis}
\label{sec:convergence}
% \textcolor{red}{TODO by SRD}
The convergence properties of the proposed method are analyzed by following the online learning framework \cite{kingma2014adam}. In this context, let us assume that the unknown convex cost functions are arranged as a sequence $f_1(\theta)$, $f_2(\theta)$,$...$, $f_T(\theta)$. At each iteration, the main goal is to explore a parameter $\theta_t$ to examine it over $f_t(\theta)$ with $t\in[1,T]$. The regret bound $R(T)$ is generally used in such optimization, where the sequence is not known in advance. Thus, an error is derived between the online guess parameter $f_t(\theta_t)$ and the expected parameter $f_t(\theta^*)$ from a feasible set $\chi$. The regret bound is computed as the sum of the error w.r.t. the guess parameters in all the previous iteration: 
\begin{equation}
R(T)=\sum_{t=1}^{T}{[f_t(\theta_t)-f_t(\theta^*)]},
\end{equation}
where $\theta^*=\mbox{arg }\mbox{min}_{\theta \in \chi }\sum_{t=1}^{T}{f_t(\theta)}$. 

In this context, the regret bound for the proposed \texttt{AngularGrad} is $O(\sqrt{T})$. %We provide the corresponding proof in the Appendix. 
Considering the following notations: $g_{t,i}$ $\rightarrow$ gradient for the $i^{th}$ element in the $t^{th}$ iteration, $g_{1:t,i}=[g_{1,i},g_{2,i},...,g_{t,i}] \in \mathbb{R}^t$ $\rightarrow$ gradient vector w.r.t. the $i^{th}$ parameter up to the $t^{th}$ iteration, and $\gamma \triangleq \frac{\beta_1^2}{\sqrt{\beta_2}}$. 

\begin{theorem}
\textit{Let us consider the bounded gradients for function $f_t$ (i.e., $||g_{t,\theta}||_2 \leq G$ and $||g_{t,\theta}||_{\infty} \leq G_{\infty}$) $\forall\theta \in \mathbb{R}^d$. Moreover, let the bounded distance be generated by the \texttt{AngularGrad} between any $\theta_t$ (i.e., $||\theta_n-\theta_m||_2 \leq D$ and $||\theta_n-\theta_m||_\infty \leq D_\infty$ for any $m,n\in\{1,...,T\}$). Let $\gamma \triangleq \frac{\beta_1^2}{\sqrt{\beta_2}}$, $\beta_1,\beta_2 \in [0,1)$ follows $\frac{\beta_1^2}{\sqrt{\beta_2}} < 1$, $\alpha_t=\frac{\alpha}{\sqrt{t}}$, and $\beta_{1,t}=\beta_1\lambda^{t-1},\lambda \in (0,1)$ with $\lambda \approx 1$. The following guarantee is satisfied by \texttt{AngularGrad}, $\forall T \geq 1$:} 
\begin{equation}
\begin{split}
R(T) & \leq \frac{D^2}{\alpha(1-\beta_1)}\sum_{i=1}^{d}{\sqrt{T\hat{v}_{T,i}}} 
\\&+ \frac{\alpha(1+\beta_1) G_\infty}{(1-\beta_1)\sqrt{1-\beta_2}(1-\gamma)^2}\sum_{i=1}^{d}{||g_{1:T,i}||_2} 
\\&+ \sum_{i=1}^{d}{\frac{D_{\infty}^{2}G_{\infty}\sqrt{1-\beta_2}}{2\alpha (1-\beta_1)(1-\lambda)^2}}
\end{split}
\end{equation}
\end{theorem}

%-------------------------------------------------

\begin{proof}[Proof]
Based on Lemma \textcolor{blue}{9.2} of Adam optimizer \cite{kingma2014adam}, the following statement can be made:
%%%%
\begin{equation}
    f_t(\theta_t)-f_t(\theta^*) \leq g_t^T(\theta_t-\theta^*) = \sum_{i=1}^{d}{g_{t,i}(\theta_{t,i}-\theta_{,i}^*)}
\label{eq:proof1}
\end{equation}
%%%%
Moreover, the following equation can be obtained by exploiting the update rule of the proposed \texttt{AngularGrad}, as mentioned in the Algorithm \ref{alg:prop} (note that we ignore the $\epsilon$ term to simplify the mathematical expression), 
\begin{dmath}
 \theta_{t+1} =\theta_t - \frac{\alpha_t {\phi_t} \hat{m}_t}{\sqrt[]{\hat{v}_{t}}}
 =\theta_t - \frac{\alpha_t {\phi_t}}{\left(1-\beta_1^t\right)} \left(\frac{\beta_{1,t}}{\sqrt[]{\hat{v}_{t}}}m_{t-1} + \frac{(1-\beta_{1,t})}{\sqrt[]{\hat{v}_{t}}}g_t\right),
\end{dmath}
where the $1^{st}$ order moment at $t^{th}$ iteration is represented by $\beta_{1,t}$.
The following can be represented for the $i^{th}$ parameter of vector $\theta_t \in R^d$:
\begin{dmath}
(\theta_{t+1,i}-\theta_{,i}^*)^2=(\theta_{t,i}-\theta_{,i}^*)^2-\frac{2\alpha_t {\phi}_{t,i}}{1-\beta_1^t}
\Big(\frac{\beta_{1,t}}{\sqrt[]{\hat{v}_{t,i}}}m_{t-1,i} + \frac{(1-\beta_{1,t})}{\sqrt[]{\hat{v}_{t,i}}}g_{t,i}\Big)(\theta_{t,i}-\theta_{,i}^*)
+\alpha_t^2 {\phi}_{t,i}^2 (\frac{\hat{m}_{t,i}}{\sqrt{\hat{v}_{t,i}}})^2
\label{eq:proof2}
\end{dmath}
Consequently, Eq. (\ref{eq:proof2}) can be reordered as:
\begin{dmath}
g_{t,i}(\theta_{t,i}-\theta_{,i}^*)=\frac{(1-\beta_1^t)\sqrt{\hat{v}_{t,i}}}{2\alpha_t {\phi}_{t,i}(1-\beta_{1,t})}
\Big((\theta_{t,i}-\theta_{,i}^*)^2-(\theta_{t+1,i}-\theta_{,i}^*)^2\Big)
+\frac{\beta_{1,t}}{1-\beta_{1,t}}(\theta_{,i}^*-\theta_{t,i})m_{t-1,i}
+\frac{\alpha_t(1-\beta_1^t) {\phi}_{t,i}}{2(1-\beta_{1,t})}\frac{(\hat{m}_{t,i})^2}{\sqrt{\hat{v}_{t,i}}}.
\end{dmath}
Further, it can be rewritten as:
\begin{dmath}
g_{t,i}(\theta_{t,i}-\theta_{,i}^*) =\frac{(1-\beta_1^t)\sqrt{\hat{v}_{t,i}}}{2\alpha_t {\phi}_{t,i}(1-\beta_{1,t})}
\Big((\theta_{t,i}-\theta_{,i}^*)^2-(\theta_{t+1,i}-\theta_{,i}^*)^2\Big)
 +\sqrt{\frac{\beta_{1,t}}{\alpha_{t-1}(1-\beta_{1,t})}(\theta_{,i}^*-\theta_{t,i})^2\sqrt{\hat{v}_{t-1,i}}} \sqrt{\frac{\beta_{1,t}\alpha_{t-1}(m_{t-1,i})^2}{(1-\beta_{1,t})\sqrt{\hat{v}_{t-1,i}}}}
+\frac{\alpha_t(1-\beta_1^t){\phi}_{t,i}}{2(1-\beta_{1,t})}\frac{(\hat{m}_{t,i})^2}{\sqrt{\hat{v}_{t,i}}}
\end{dmath}
We can further reorganize by utilizing the Young's inequality, i.e. $ab \leq a^2/2+b^2/2$, as well as the property that $\beta_{1,t} \leq \beta_1$ as:
\begin{dmath}
g_{t,i}(\theta_{t,i}-\theta_{,i}^*) \leq \frac{1}{2\alpha_t{\phi}_{t,i}(1-\beta_1)}
\Big((\theta_{t,i}-\theta_{,i}^*)^2-(\theta_{t+1,i}-\theta_{,i}^*)^2\Big)\sqrt{\hat{v}_{t,i}}
 +\frac{\beta_{1,t}}{2\alpha_{t-1}(1-\beta_{1,t})}(\theta_{,i}^*-\theta_{t,i})^2\sqrt{\hat{v}_{t-1,i}} + \frac{\beta_1\alpha_{t-1}(m_{t-1,i})^2}{2(1-\beta_1)\sqrt{\hat{v}_{t-1,i}}}
+\frac{\alpha_t{\phi}_{t,i}}{2(1-\beta_1)}\frac{(\hat{m}_{t,i})^2}{\sqrt{\hat{v}_{t,i}}}
\end{dmath}
It is evident that the minimum and maximum values of angular coefficient (${\phi}$) are $0.5$ and $1$, respectively. Thus, $0.5 \leq {\phi}_{t,i} \leq 1$. We can remove ${\phi}_{t,i}$ from last term of above equation without loss of the inequality property. Therefore, we can rewrite it as:
\begin{dmath}
g_{t,i}(\theta_{t,i}-\theta_{,i}^*) \leq \frac{1}{2\alpha_t{\phi}_{t,i}(1-\beta_1)}
\Big((\theta_{t,i}-\theta_{,i}^*)^2-(\theta_{t+1,i}-\theta_{,i}^*)^2\Big)\sqrt{\hat{v}_{t,i}}
 +\frac{\beta_{1,t}}{2\alpha_{t-1}(1-\beta_{1,t})}(\theta_{,i}^*-\theta_{t,i})^2\sqrt{\hat{v}_{t-1,i}} + \frac{\beta_1\alpha_{t-1}(m_{t-1,i})^2}{2(1-\beta_1)\sqrt{\hat{v}_{t-1,i}}}
+\frac{\alpha_t}{2(1-\beta_1)}\frac{(\hat{m}_{t,i})^2}{\sqrt{\hat{v}_{t,i}}}
\end{dmath}
We compute the regret bound by summing it over all the parameter dimensions (i.e., $i\in \{1,\dots,d\}$) and across all the convex function sequences (i.e., $t\in \{1,\dots,T\}$). We utilize the Lemma 10.4 of Adam \cite{kingma2014adam} to aggregate the regret bound in the upper bound of $f_t(\theta_t)-f_t(\theta^*)$ as:
\begin{dmath}
R(T) \leq \sum_{i=1}^{d}{\frac{1}{2\alpha_1{\phi}_{1,i}(1-\beta_1)}} (\theta_{1,i}-\theta_{,i}^*)^2\sqrt{\hat{v}_{1,i}} + \sum_{i=1}^{d}{\sum_{t=2}^{T}{\frac{1}{2(1-\beta_1)}} (\theta_{t,i}-\theta_{,i}^*)^2(\frac{\sqrt{\hat{v}_{t,i}}}{\alpha_t{\phi}_{t,i}}-\frac{\sqrt{\hat{v}_{t-1,i}}}{\alpha_{t-1}{\phi}_{t-1,i}})}
 + \frac{\beta_1\alpha G_\infty}{(1-\beta_1)\sqrt{1-\beta_2}(1-\gamma)^2}\sum_{i=1}^{d}{||g_{1:T,i}||_2}
+ \frac{\alpha G_\infty}{(1-\beta_1)\sqrt{1-\beta_2}(1-\gamma)^2}\sum_{i=1}^{d}{||g_{1:T,i}||_2}
 + \sum_{i=1}^{d}{\sum_{t=1}^{T}{\frac{\beta_{1,t}}{2\alpha_{t}(1-\beta_{1,t})}(\theta_{,i}^*-\theta_{t,i})^2\sqrt{\hat{v}_{t,i}}}}
\end{dmath}

We incorporate the assumptions that $\alpha=\alpha_t\sqrt{t}$, $||\theta_t-\theta^*||_2 \leq D$ and $||\theta_m-\theta_n||_{\infty} \leq D_{\infty}$, the above equation is refined as:
\begin{dmath}
R(T) \leq \frac{D^2}{2\alpha(1-\beta_1)}\sum_{i=1}^{d}{\frac{\sqrt{T\hat{v}_{T,i}}}{{\phi}_{1,i}}} 
+ \frac{\alpha(1+\beta_1) G_\infty}{(1-\beta_1)\sqrt{1-\beta_2}(1-\gamma)^2}\sum_{i=1}^{d}{||g_{1:T,i}||_2}
+ \frac{D_{\infty}^{2}}{2\alpha}\sum_{i=1}^{d}{\sum_{t=1}^{t}{\frac{\beta_{1,t}}{(1-\beta_{1,t})}\sqrt{t\hat{v}_{t,i}}}}
  \leq \frac{D^2}{2\alpha(1-\beta_1)}\sum_{i=1}^{d}{\frac{\sqrt{T\hat{v}_{T,i}}}{{\phi}_{1,i}}} 
+ \frac{\alpha(1+\beta_1) G_\infty}{(1-\beta_1)\sqrt{1-\beta_2}(1-\gamma)^2}\sum_{i=1}^{d}{||g_{1:T,i}||_2} 
+ \frac{D_{\infty}^{2}G_{\infty}\sqrt{1-\beta_2}}{2\alpha}\sum_{i=1}^{d}{\sum_{t=1}^{t}{\frac{\beta_{1,t}}{(1-\beta_{1,t})}\sqrt{t}}}
\end{dmath}

\begin{figure*}[!ht]
\centering
\includegraphics[width=0.9\linewidth]{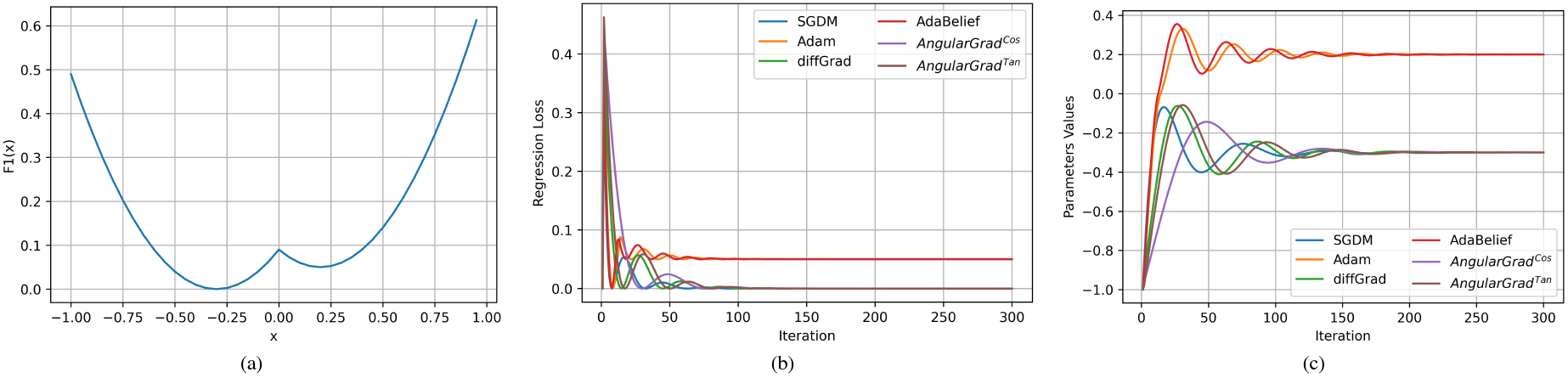}
\includegraphics[width=0.9\linewidth]{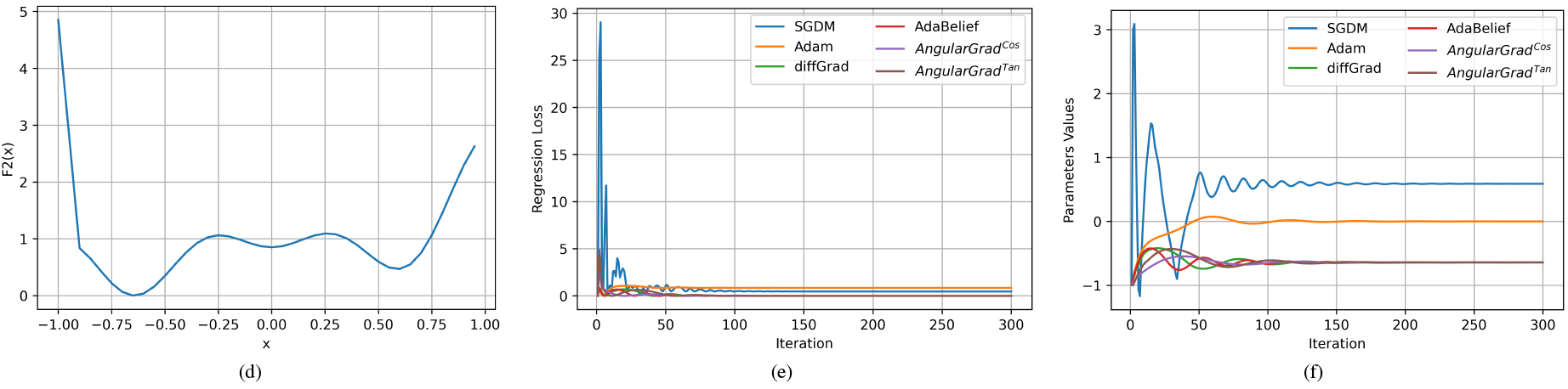}
\includegraphics[width=0.9\linewidth]{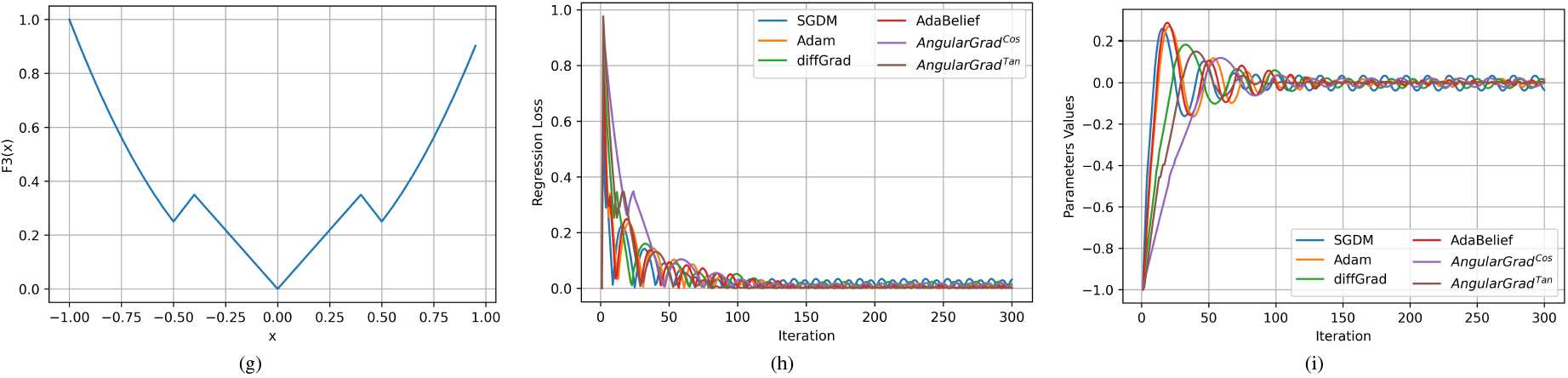}
\caption{Empirical results over toy examples using SGDM, Adam, diffGrad, AdaBelief, \texttt{AngularGrad$^{\cos}$} and \texttt{AngularGrad$^{\tan}$}.}
\label{fig:analytical}
\end{figure*}

By utilizing the following finding of Adam \cite{kingma2014adam}: $\sum_{t=1}^{t}{\frac{\beta_{1,t}}{(1-\beta_{1,t})}\sqrt{t}} \leq \frac{1}{(1-\beta_1)(1-\gamma)^2}$, the regret bound can be further refined as:
\begin{dmath}
R(T) \leq \frac{D^2}{2\alpha(1-\beta_1)}\sum_{i=1}^{d}{\frac{\sqrt{T\hat{v}_{T,i}}}{{\phi}_{1,i}}} 
+ \frac{\alpha(1+\beta_1) G_\infty}{(1-\beta_1)\sqrt{1-\beta_2}(1-\gamma)^2}\sum_{i=1}^{d}{||g_{1:T,i}||_2} 
+ \sum_{i=1}^{d}{\frac{D_{\infty}^{2}G_{\infty}\sqrt{1-\beta_2}}{2\alpha (1-\beta_1)(1-\lambda)^2}}
\end{dmath}

Referring to the proposed \texttt{AngularGrad} presented in Algorithm \ref{alg:prop} and considering $\lambda_1 = \lambda_2 = 0.5$, we need to compute the bound for angular coefficient ${\phi}_{1,i}$. We know that $0 \leq tanh(|.|) \leq 1$. Thus, $0.5 \leq {\phi}_{1,i} \leq 1$. Furthermore, the regret bound for the proposed \texttt{AngularGrad} can be formulated as:
\begin{dmath}
R(T) \leq \frac{D^2}{2\alpha(1-\beta_1)}\sum_{i=1}^{d}{\frac{\sqrt{T\hat{v}_{T,i}}}{0.5}} 
+ \frac{\alpha(1+\beta_1) G_\infty}{(1-\beta_1)\sqrt{1-\beta_2}(1-\gamma)^2}\sum_{i=1}^{d}{||g_{1:T,i}||_2} 
+ \sum_{i=1}^{d}{\frac{D_{\infty}^{2}G_{\infty}\sqrt{1-\beta_2}}{2\alpha (1-\beta_1)(1-\lambda)^2}}
\end{dmath}

Finally, It can be written following Eq. (\ref{eq:proofFinal}):  
\begin{dmath}
R(T) \leq \frac{D^2}{\alpha(1-\beta_1)}\sum_{i=1}^{d}{\sqrt{T\hat{v}_{T,i}}} 
+ \frac{\alpha(1+\beta_1) G_\infty}{(1-\beta_1)\sqrt{1-\beta_2}(1-\gamma)^2}\sum_{i=1}^{d}{||g_{1:T,i}||_2} 
+ \sum_{i=1}^{d}{\frac{D_{\infty}^{2}G_{\infty}\sqrt{1-\beta_2}}{2\alpha (1-\beta_1)(1-\lambda)^2}}
\label{eq:proofFinal}
\end{dmath}
\end{proof}

%-------------------------------------------------
It is evident that the aggregating term over dimension $d$ can be significantly lower than the corresponding upper bound. Therefore, $\sum_{i=1}^{d}{||g_{1:T,i}||_2}<< dG_\infty\sqrt{T}$ and $\sum_{i=1}^{d}{\sqrt{T\hat{v}_{T,i}}} << dG_\infty\sqrt{T}$. Mostly, adaptive methods such as Adam or the proposed \texttt{AngularGrad} achieve $O(\log d\sqrt{T})$, which is better than $O(\sqrt{dT})$ achieved by non-adaptive methods. Similar to Adam \cite{kingma2014adam}, the proposed \texttt{AngularGrad} optimizer also utilizes the decay of $\beta_{1,t}$ to conduct the theoretical analysis.

Finally, considering the above theorem coupled with $\sum_{i=1}^{d}{||g_{1:T,i}||_2}<< dG_\infty\sqrt{T}$ and $\sum_{i=1}^{d}{\sqrt{T\hat{v}_{T,i}}} << dG_\infty\sqrt{T}$, we show the average regret convergence of the proposed \texttt{AngularGrad} in a corollary below.
\begin{corollary}
\textit{Let us consider the bounded gradients for function $f_t$, i.e. $||g_{t,\theta}||_2 \leq G$ and $||g_{t,\theta}||_{\infty} \leq G_{\infty}$, $\forall\theta \in \mathbb{R}^d$. The bounded distance is generated by the \texttt{AngularGrad} between any $\theta_t$, i.e., $||\theta_n-\theta_m||_2 \leq D$ and $||\theta_n-\theta_m||_\infty \leq D_\infty$, $\forall m,n\in\{1,...,T\}$. The following guarantee is shown by \texttt{AngularGrad} $\forall T \geq 1$:} 
\begin{equation}
\frac{R(T)}{T}=O(\frac{1}{\sqrt{T}}). 
\end{equation}
Thus, $\lim_{T\rightarrow\infty}\frac{R(T)}{T}=0$.
\end{corollary}

\section{Empirical Analysis}
\label{sec:empirical}
The diffGrad~\cite{dubey2019diffgrad} optimizer already introduced the difference of gradient-based friction concept for optimization purposes. In order to achieve sharp changes in convergence, we leap another step forward and exploit the benefits of using the angle between two consecutive gradients. To justify our theory, we model the optimization problem as a regression one over three one-dimensional non-convex functions, performing optimization over these functions using SGDM, Adam, diffGrad, AdaBelief, \texttt{AngularGrad$^{\cos}$} and \texttt{AngularGrad$^{\tan}$}.

The three non-convex functions (considered from diffGrad \cite{dubey2019diffgrad}), namely $F1$, $F2$, $F3$, are as follows:
\begin{equation}
    F1(x)=
    \begin{cases}
    (x+0.3)^{2}, & {\rm for}~x\leq0 \\
    (x-0.2)^{2}+0.05, & {\rm for}~ x>0 \\
    \end{cases}
\end{equation}
\begin{equation}
    F2(x)=
    \begin{cases}
    -40x-35.15, & {\rm for}~ x\leq-0.9 \\
    x^{3}+xsin(8x)+0.85 & {\rm for}~ x>-0.9
    \end{cases}
\end{equation}
\begin{equation}
    F3(x)=
    \begin{cases}
    x^{2}, & {\rm for}~x\leq-0.5\\
    0.75+x, & {\rm for}~-0.5<x\leq-0.4\\
    -7x/8, & {\rm for}~-0.4<x\leq0\\
    7x/8, & {\rm for}~0<x\leq0.4\\
    0.75-x, & {\rm for}~0.4<x\leq0.5\\
    x^{2}, & {\rm for}~0.5<x
    \end{cases}
\end{equation}

where $x\in(-\infty,+\infty)$ is the input. Fig. \ref{fig:analytical} depicts the functions $F1$ (first row), $F2$ (second row), and $F3$ (third row) as $x\in(-1,+1)$. The function $F1$ has one local minimum, while $F2$ and $F3$ have two local minima. All the optimizers in the experiment share the following settings: the decay rates of first and second order moments $(\beta_{1},\beta_{2})$ are 0.95 and 0.999; the first and second order moments ($m,v$) are initialized to 0; the learning rate $\alpha$ is 0.1, and the parameter $\theta$ is initialized to -1. The previous gradient for the $1^{st}$ step ($g_{0}$) is 0. All the optimizers are run for $300$ iterations. The regression loss and the parameter value $\theta$ are recorded for the analysis.
%%%%%%%%%%%%%%%%%%%%%%%%%%%%%%%%%%%%%
%%%%%%%%%%% ROSENBROCK %%%%%%%%%%%%%%
%%%%%%%%%%%%%%%%%%%%%%%%%%%%%%%%%%%%%
\begin{figure*}[!ht]
\centering
\resizebox{\linewidth}{!}{
\begin{tabular}{cccc}
\includegraphics[trim={40 40 40 40}, clip, width=\linewidth]{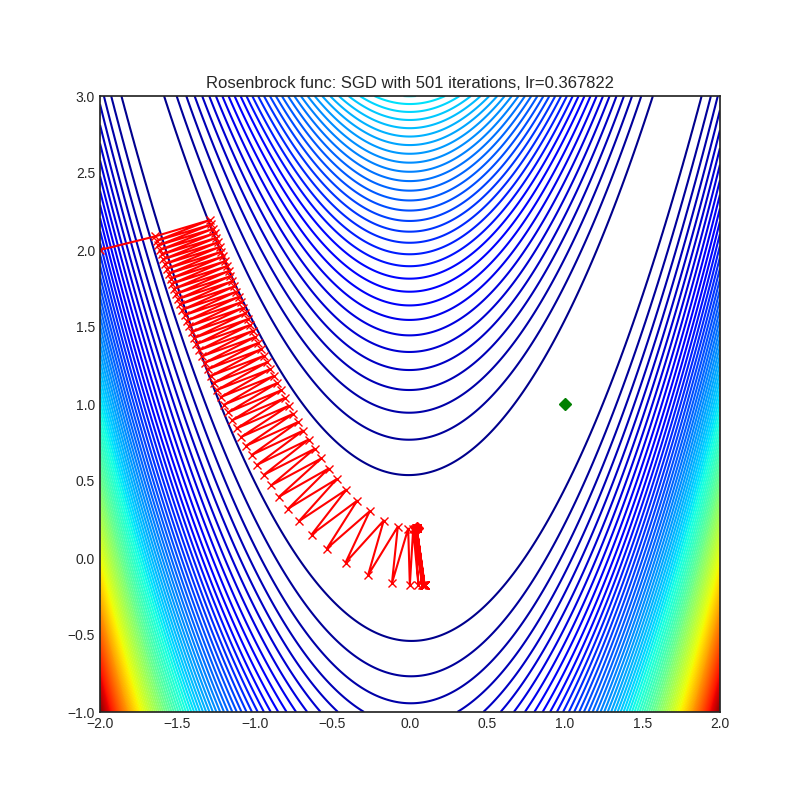} &
\includegraphics[trim={40 40 40 40}, clip, width=\linewidth]{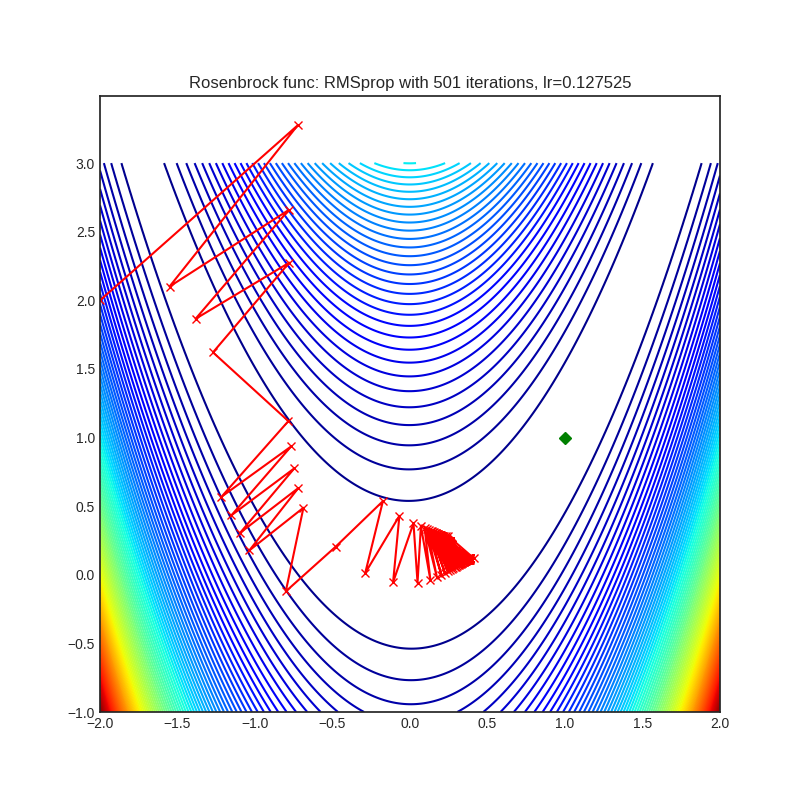} &
\includegraphics[trim={40 40 40 40}, clip, width=\linewidth]{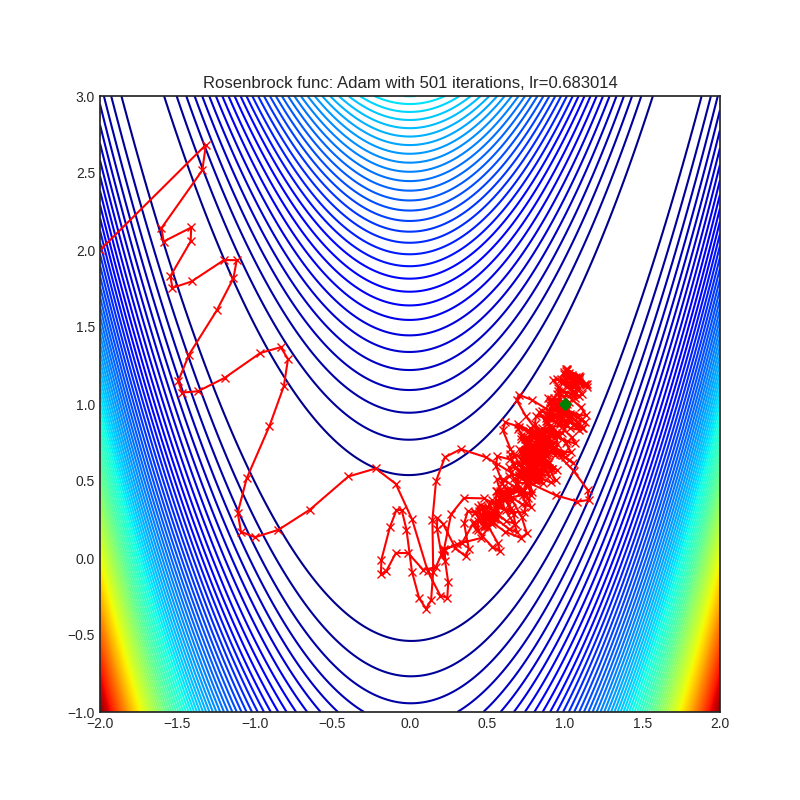} &
\includegraphics[trim={40 40 40 40}, clip, width=\linewidth]{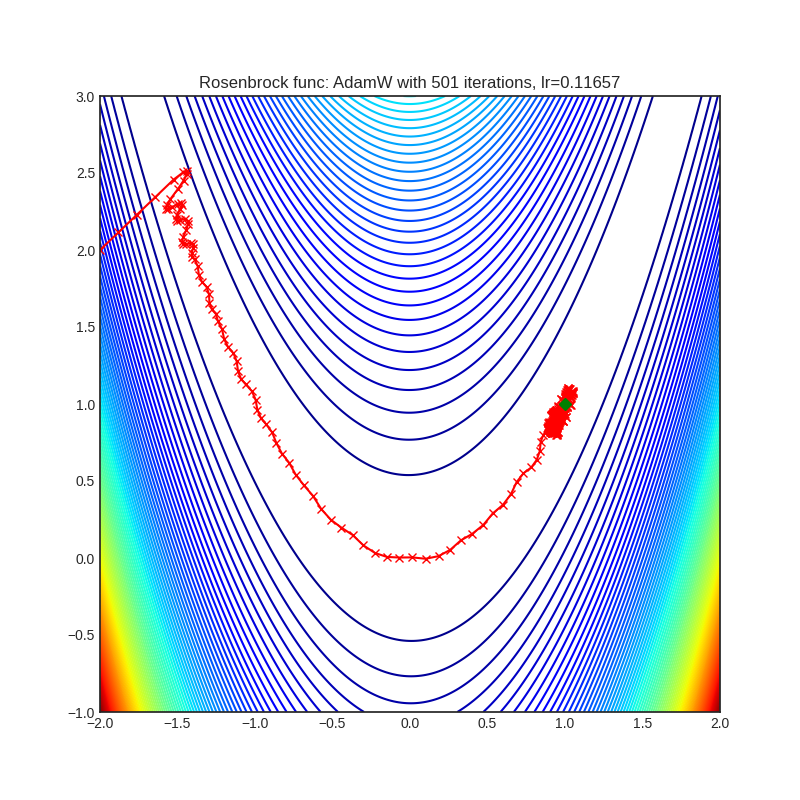} \\
\huge (SGD) & \huge (RMSprop) & \huge (Adam) & \huge (AdamW) \\
\includegraphics[trim={40 40 40 40}, clip, width=\linewidth]{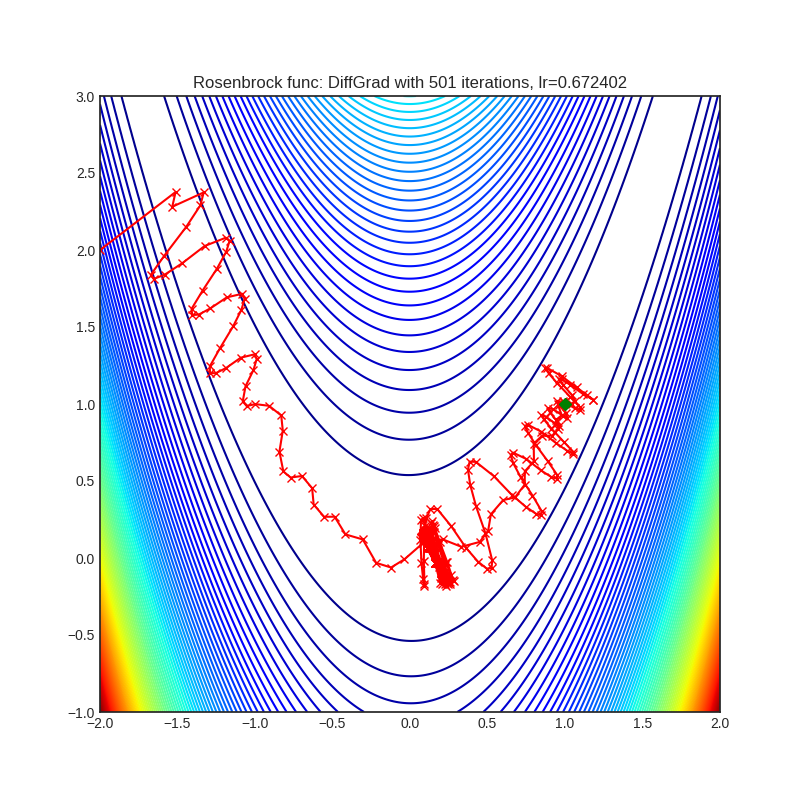} &
\includegraphics[trim={40 40 40 40}, clip, width=\linewidth]{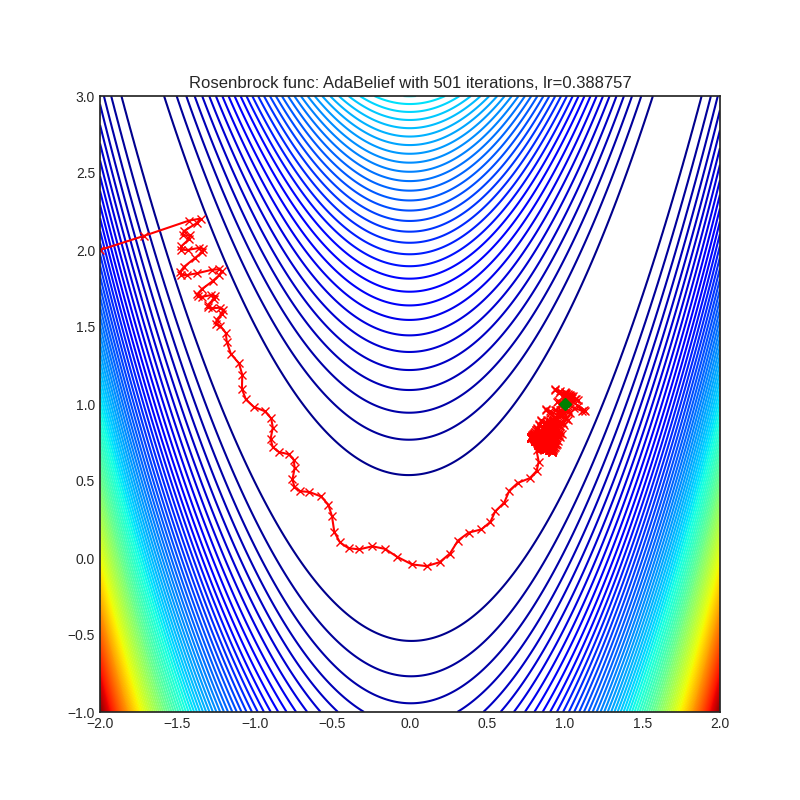} &
\includegraphics[trim={40 40 40 40}, clip, width=\linewidth]{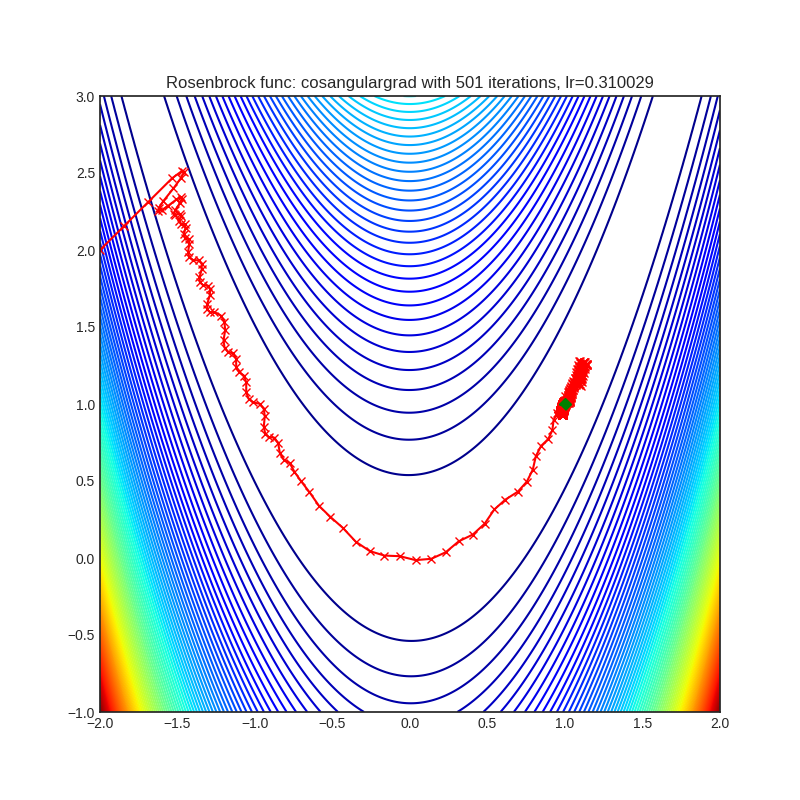} &
\includegraphics[trim={40 40 40 40}, clip, width=\linewidth]{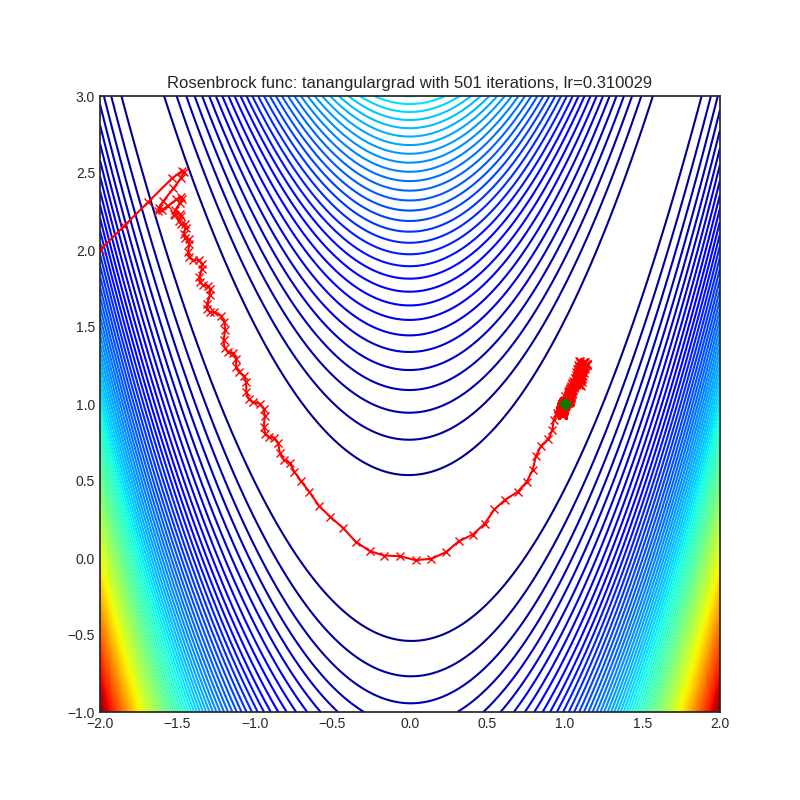} \\
\huge (diffGrad) & \huge (AdaBelief)  & \huge ($AngularGrad^{cos}$) & \huge ($AngularGrad^{tan}$)\\
\end{tabular}}
\caption{Graphical representation of gradient trajectories using different optimization algorithms (in particular, SGD, RMSprop, Adam, AdamW, diffGrad, AdaBelief \texttt{AngularGrad$^{\cos}$} and \texttt{AngularGrad$^{\tan}$}) by considering the Rosenbrock function. These graphics have been obtained from \url{https://github.com/jettify/pytorch-optimizer} with $seed=2$}
\label{fig:analytical12}
\end{figure*}
%%%%%%%%%%%%%%%%%%%%%%%%%%%%%%%%%%%%%%

Moreover, the first column provides the function shapes, while the second and third columns indicate the regression loss vs. the number of iterations and the number of parameters vs. the number of iterations, respectively. For F1(x), Fig. \ref{fig:analytical}(c) clearly shows that Adam and AdaBelief overshoots the global minimum at $\theta = -0.3$ due to the high momentum gained, and finally converges at $\theta = 0.2$. Neither of them are able to accumulate the zero loss, which is resolved in the case of SGDM, diffGrad, \texttt{AngularGrad$^{\cos}$} and \texttt{AngularGrad$^{\tan}$}. The justification brings forward the introduction of the angular coefficient $\phi$ and its advantages, as described by the theorems presented in previous sections in the sense of reducing the zigzagging nature (i.e., the noise) of the curve while reaching towards the global minimum. Hence, it avoids overshooting. A similar behavior is also depicted in Figs. \ref{fig:analytical}(e) and \ref{fig:analytical}(f) for F2(x) where not only Adam but also SGDM are not able to accumulate zero loss. Finally, in Figs. \ref{fig:analytical}(h) and \ref{fig:analytical}(i) for F3(x), we observe that all the optimizers lead to nearly zero loss and are successfully able to avoid being trapped in local minima and reach the global minimum. However, the oscillations of \texttt{AngularGrad$^{\cos}$} and \texttt{AngularGrad$^{\tan}$} are smaller near the global minimum. Thus, they achieve a more precise convergence. In summary, the empirical analysis clearly reveals that, among the aforementioned functions, \texttt{AngularGrad$^{\cos}$} and \texttt{AngularGrad$^{\tan}$} do not get stuck into local minima and converge to the global minimum much faster when compared to other competing optimizers.

%clearly shows that Adam, AdaBelief and \texttt{AngularGrad$^{\cos}$} overshoot the global minimum at $\theta = -0.3$ due to the high momentum gained and finally converges at $\theta = 0.2$. Neither they are able to accumulate the zero loss which is resolved in case of diffGrad and \texttt{AngularGrad$^{\tan}$}. The justification brings forward the introduction of the angular coefficient $\phi$ and its advantages described by the theorems (presented in previous sections) which diminishes the zigzagging nature, i.e., the noise of the curve while reaching towards global minimum. Hence, it avoids overshooting. In Figs. \ref{fig:analytical}(e) and \ref{fig:analytical}(f), considering F2(x), we observe that Adam overshoot the global minimum while the other optimizers are able to reach it, hence accumulating zero loss. Finally, in Figs. \ref{fig:analytical}(h) and \ref{fig:analytical}(i) for F3(x), we observe that all the optimizers give nearly zero loss and are successfully able to cross local minima and reach the global minimum. However, the oscillations of \texttt{AngularGrad$^{\cos}$} and \texttt{AngularGrad$^{\tan}$} are smaller near the global minimum. Thus, they lead to a more precise convergence.

\section{Experimental Results}
\label{sec:experiments}

\subsection{Experimental Settings}
%%%%%%
Extensive experiments have been conducted in order to test the behavior and validate the performance of the newly proposed \texttt{AngularGrad} optimizer. The conducted experiments are briefly summarized below:
%%%%%%%%%%%%%%%%%%%
\begin{itemize}
%%%%
    \item The first experiment studies and analyzes the behaviour of the proposed optimization algorithm when dealing with a challenging benchmark function, with the aim of to understanding the potentials of the proposed optimization method. In this sense, the proposed algorithm has been evaluated on {Rosenbrock} function.
    %%%%
    \item The second experiment tackles a classification task, evaluating the performance of the proposed optimizer over several CNNs and compares the obtained results in terms of overall accuracy (OA) with other well-known optimizers. In this sense, CIFAR10 and CIFAR100 \cite{krizhevsky2009learning} data sets have been taken into account.
    %%%%
    \item The third experiment delves into the problem of classification, comparing the performance of our optimizer against some other widely-used optimizers on a challenging Mini-ImageNet data set \cite{vinyals2016matching}. In this sense, six different deep architectures have been tested, in particular the ResNet18, ResNet50 and ResNet101 models as well as their weight-standardised counterparts \cite{qiao2019micro}, i.e. ResNet18ws, ResNet50ws and ResNet101ws, respectively.
    %%%%
    \item The fourth experiment extends the classification problem, testing the proposed optimizer over the popular classification benchmark of ImageNet \cite{krizhevsky2009learning}, drawing a comparison between the proposed \texttt{AngularGrad} and the results obtained by others optimizers such as AdaBelief and MSV AG \cite{zhuang2020adabelief}, SGDM, AdaBound, Yogi, Adam, and AdamW \cite{chen2018closing}, and RAdam \cite{liu2019radam}.
    %%%%
    \item The fifth experiment explores the performance of the proposed method in the task of fine-grained classification, evaluating its skills when distinguishing subordinate categories within entry level categories. Four fine-grained image classification data sets have been considered (Stanford Cars \cite{krause20133d}, Stanford Dogs \cite{khosla2011novel}, FGVC Aircraft \cite{maji2013fine} and CUB-200-2011 \cite{wah2011caltech}), with different optimizers using ResNet50 model for comparison purposes.
    %%%
    \item The sixth experiment visualizes the effects of using different optimizers on the network architecture, using the ResNet50 model as a baseline and the Mini-Imagenet data set. Specifically, we visualize the 3D loss landscape trajectory resulting from different optimizers to analyze uniformity.
\end{itemize}

\begin{table*}[!ht]
\let\center\empty
\let\endcenter\relax
\centering
\caption{Classification results over the CIFAR10 data set. Best results are highlighted in bold, whilst second best results are in blue.}
\resizebox{0.9\linewidth}{!}{\input{tables/cifar10.tex}}
\label{table:compCIFAR10}
\end{table*}

\begin{figure*}[!ht]
\centering
\includegraphics[width=\linewidth]{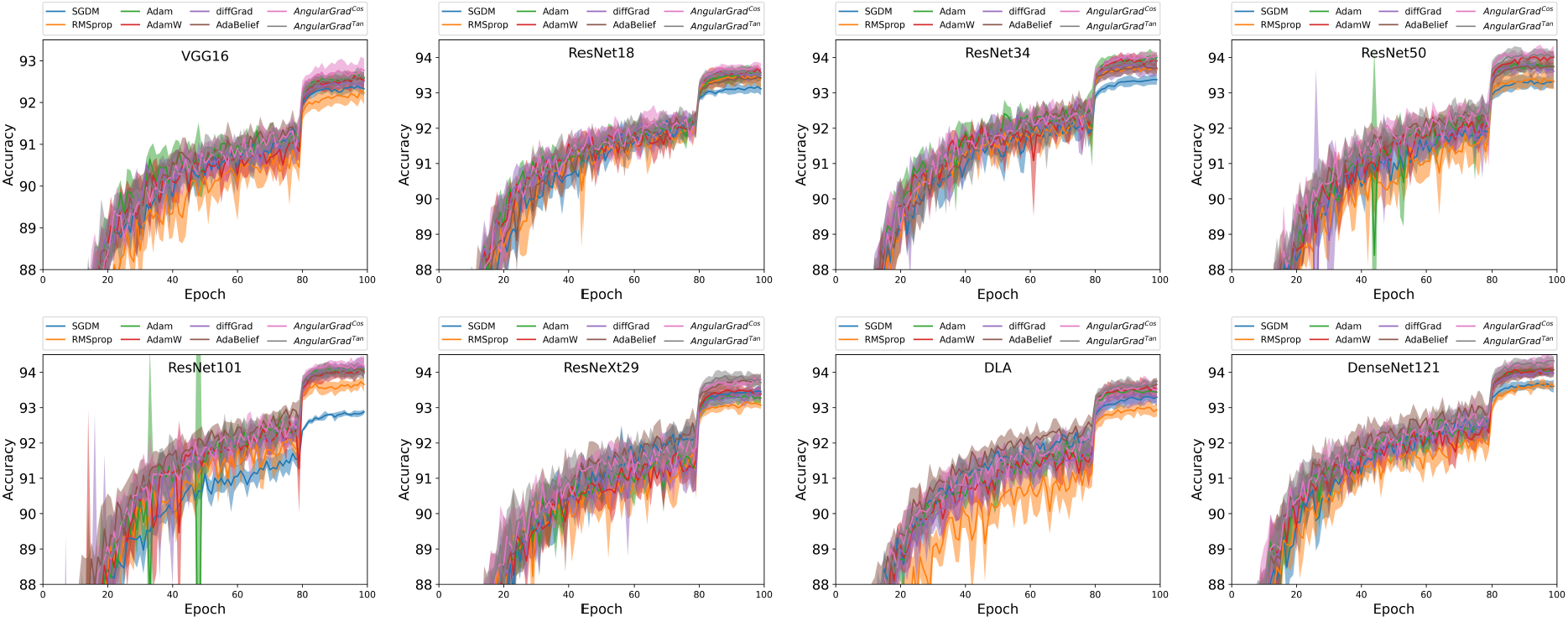}
\caption{Classification overall accuracy (OA) over the CIFAR10 data set using different optimizers with VGG16, ResNet18, ResNet34, ResNet50, ResNet101, DLA, ResNeXt29 and DenseNet121 models.} 
\label{fig:CIFAR10}
\end{figure*}

\begin{table*}[!ht]
\let\center\empty
\let\endcenter\relax
\centering
\caption{Classification results over the CIFAR100 data set. Best results are highlighted in bold, whilst second best results are in blue.}
\resizebox{\linewidth}{!}{\input{tables/cifar100.tex}}
\label{table:compCIFAR100}
\end{table*}

\begin{figure*}[!ht]
\centering
\includegraphics[width=\linewidth]{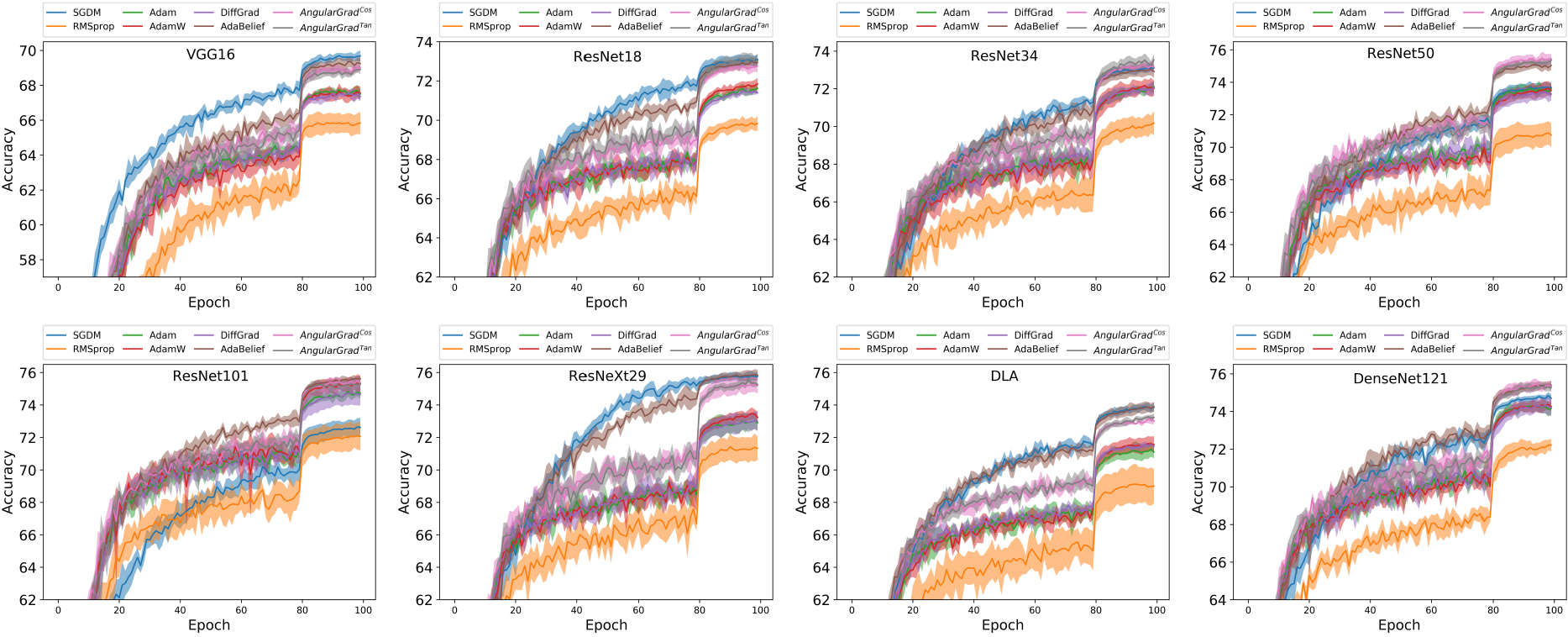}
\caption{Classification overall accuracy (OA) over the CIFAR100 data set using different optimizers with VGG16, ResNet18, ResNet34, ResNet50, ResNet101, ResNeXt29, DLA and DenseNet121 models.}
\label{fig:CIFAR100}
\end{figure*}

\subsection{Results and Analysis}
%%%%%%
% The extensive comparisons are performed with different optimizers including SGDM \cite{sutskever2013importance}, RMSprop \cite{hinton2012neural}, Adam \cite{kingma2014adam}, AdamW \cite{loshchilov2017decoupled}, diffGrad \cite{dubey2019diffgrad}, \texttt{AngularGrad$^\cos$}, and \texttt{AngularGrad$^\tan$} on three different classification tasks. 

%evaluated using seven popular and well-known deep models i.e., VGG11 \cite{simonyan2014very}, ResNet18 \cite{he2016deep}, ResNet34 \cite{he2016deep}, ResNet50 \cite{he2016deep}, ResNet101 \cite{he2016deep}, ResNeXt29 \cite{xie2017aggregated} and DenseNet121 \cite{huang2017densely}.
\subsubsection{Convergence analysis on Rosenbrock function}
%%%%%%%%%%%%
\textbf{Rosenbrock} function (also called Valley function) \cite{rosenbrock1960automatic} is a widely used unimodal-problem for testing gradient-based optimization algorithms. The continuous $N$-dimensional Rosenbrock function contains only one global minimum, which lies in a narrow-parabolic valley defined as follows:
%%%%%%%%%%%%%%%%%%%%
\begin{equation}
\begin{split}
   f(\mathbf{x}) = 
               f(x_1,\dots,x_N) =\\
               \sum_{i=1}^{N-1} 
               \left[ 
               b \left( x_{i+1} - x_i^2 \right)^2 +
               \left( a - x_i \right)^2
               \right] 
\end{split}
%f(x,y)=(a-x)^2+b(y-x^2)^2
\label{eq:rosen}
\end{equation}
%%%%%%%%%%%%%%%%%%%%%
where $\mathbf{x}=[x_1, x_2,\dots, x_N]\in\mathbb{R}^N$ defines the inputs with the positive integer $N$, while $a$ and $b$ are two constant hyperparameters which are usually set to $a=1$ and $b=100$. Moreover, the function is usually evaluated on the hypercubes $x_i\in[-5, 10]$ and $x_i\in[-2.048, 2.048]$, $\forall i\in[1,N]$. In this sense, although finding the valley is quite simple, converging to the global minimum is pretty difficult when $a\neq 0$ (trivial case, where the function is symmetric). As we can observe, this function has a global minimum at $f(\mathbf{x}')=0$, where $\mathbf{x}'=(1,\dots,1)\in\mathbb{R}^N$. %$(x,y)=(a,a^2)$, where $f(x,y)=0$.

To evaluate the convergence path, we have tested both the proposed \texttt{AngularGrad} optimizer along with different widely used optimization algorithms using the popular benchmark test function for comparative purposes. Particularly, SGD, RMSprop, Adam, AdamW, diffGrad, AdaBelief \texttt{AngularGrad$^{\cos}$} and \texttt{AngularGrad$^{\tan}$} have been considered.

In this sense, Fig.~\ref{fig:analytical12} provides the obtained results for Rosenbrock function. As we can observe, SGD and RMSprop are unable to reach the global minimum, while their path follows a strong and unparalleled zigzagging trajectory. The remainder algorithms are able to reach the global minimum. Notwithstanding, the trajectories of Adam and diffGrad are very noisy again. In this regard, AdamW, $AngularGrad^{tan}$ and $AngularGrad^{cos}$ have relatively smooth and uniform trajectory, reaching a solution near to the global minimum, while AdaBelief is noisier than other three optimization methods.

\subsubsection{Classification results on CIFAR10 and CIFAR100 data sets}
To conduct the image classification task on CIFAR10 and CIFAR100, the performance of the proposed optimizer is evaluated with eight deep CNN models, i.e., VGG16 \cite{simonyan2014very}, ResNet18 \cite{he2016deep}, ResNet34 \cite{he2016deep}, ResNet50 \cite{he2016deep}, ResNet101 \cite{he2016deep}, ResNeXt29 \cite{xie2017aggregated}, DLA \cite{yu2018deep} and DenseNet121 \cite{huang2017densely}. The results are compared with several optimization methods\footnote{
%%%% FOOT NOTE STARTS
In particular, SGDM has been implemented with learning rate $\mu=0.01$ and momentum $\gamma=0.9$, RMSprop with $\mu=1e-3$, smoothing constant $\alpha=0.99$ and numerical stabilizer $\epsilon=1e-8$, while the other optimization algorithms have been implemented with $\mu=1e-3$, $\beta_1=0.9$, $\beta_2=0.999$, $\epsilon=1e-8$. All of them have weight decay $WD=0.0$. Furthermore, the learning rate scheduler divides the learning rate by 10 in the 80th epoch.}.
%%%%% FOOT NOTE ENDS
Furthermore, data augmentation has been applied\footnote{
%%%%%% FOOT NOTE STARTS
A padding of 4 is added to each image to obtain $40\times40$ inputs. Then, random cropping is applied to reshape the data to $32\times32$ inputs. Finally, random horizontal flip with probability of 50\% is applied.% to increase the variability.
}.
%%%% FOOT NOTE ENDS
%\textcolor{red}{Before performing the aforementioned classification task, the optimal parameters are searched for each optimizers} and comparing the obtained results in terms of mean and standard deviation
The obtained results have been compared in terms of OA, where the mean and standard deviation ($|\mu\pm\sigma|$) measurements have been obtained from 5 runs with random initialization of the network weights for all the well-known optimization methods. Tables~\ref{table:compCIFAR10} and \ref{table:compCIFAR100} evaluate the classification results on CIFAR10 and CIFAR100. Particularly, the reported results of \texttt{AngularGrad} are evaluated with $\beta_1$ = 0.9, $\beta_2$ = 0.999, $\epsilon$ = 1e-8, $WD$ = 0.0 and $\mu$=1e-3.

On the one hand, focusing on CIFAR10 dataset, Table~\ref{table:compCIFAR10} demonstrates the superiority of the proposed method in all networks evaluated. Indeed, proposed \texttt{AngularGrad$^\cos$} and \texttt{AngularGrad$^\tan$} achieve the best and second best OA result in almost all networks. On the other hand, focusing on CIFAR100 dataset, it can be observed from Table~\ref{table:compCIFAR100} that the proposed \texttt{AngularGrad} optimization algorithms generally outperforms the OA results obtained by the other optimization methods in almost all deep networks. Particularly, both \texttt{AngularGrad}-based optimizers reach the best OA in ResNet18, ResNet34, ResNet50 and DenseNet121, whilst in ResNet101 \texttt{AngularGrad$^\cos$} obtains the second best OA result. In the rest of the neural networks (i.e., VGG16, ResNeXt29 and DLA), we can observe that, although the OAs obtained by the proposed method are slightly lower, the difference between they and the best OA is very small. For instance, in VGG16, the \texttt{AngularGrad} is quite close to SGDM and AdaBelief, whilst the rest of the optimization methods (Adam, AdamW, DiffGrad and RMSprop) are at least two points behind the best OA. A similar situation is found in ResNeXt29 and DLA, where the proposed methods are pretty close to the best OA reached by AdaBelief and SGDM, respectively. 

That means the performance of \texttt{AngularGrad} algorithms are in line with the best optimization methods, matching those classification results achieved by the current state-of-the-art algorithms. Also, it is worth noting that the proposed methods improve the results of the different adaptive methods, achieving not only a comparative accuracy with diffGrad and AdamW, but outperforming it by far in most cases, for instance.

From Figs. \ref{fig:CIFAR10} and \ref{fig:CIFAR100}, we can see that the proposed methods achieve faster convergence with minimum variations ($\pm \mu$). In this sense, \texttt{AngularGrad} algorithms achieve better classification accuracy as compared to adaptive methods, such as Adam or diffGrad. Our experimental validation reveals both faster and smoother convergence and generalization power of the proposed \texttt{AngularGrad} optimizer.

\begin{figure*}[!ht]
\centering
\includegraphics[width=\linewidth]{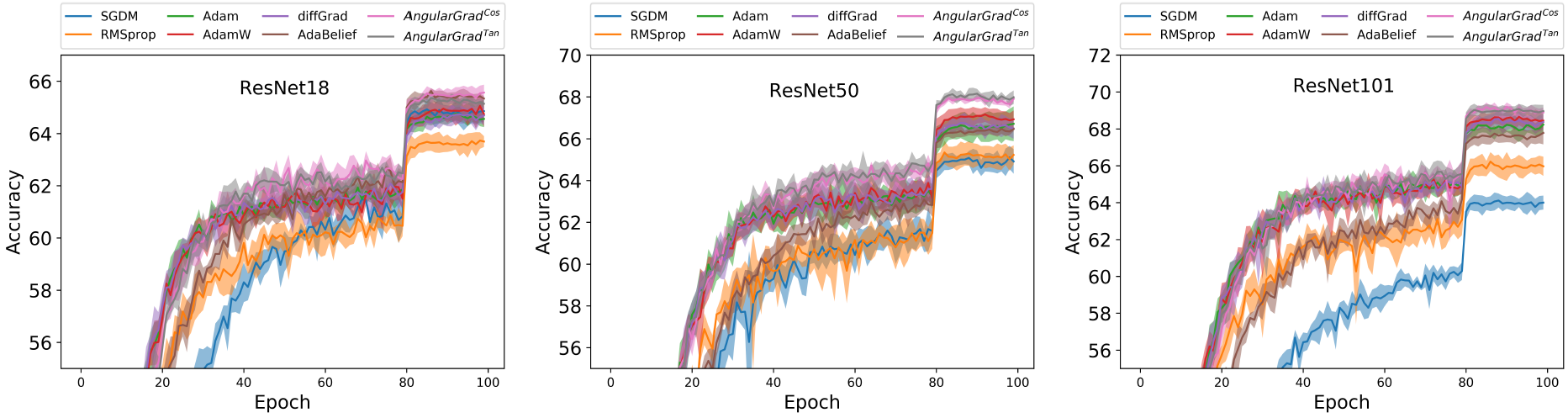}
\includegraphics[width=\linewidth]{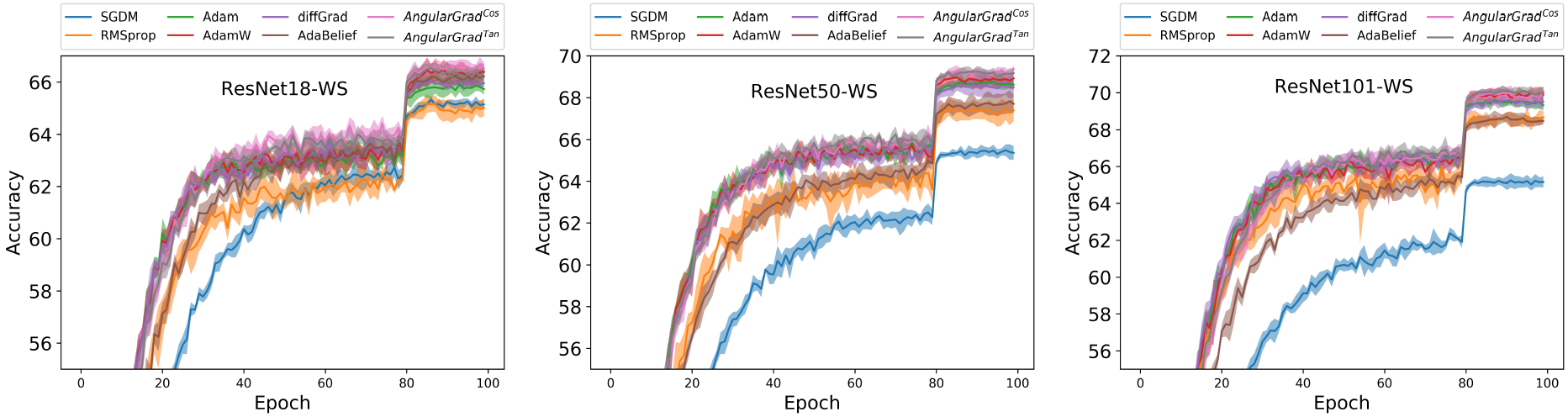}
\caption{Graphical visualization of the OA evolution achieved by the proposed optimization algorithms \texttt{AngularGrad$^\tan$} and \texttt{AngularGrad$^\cos$} on the Mini-ImageNet data set, which have been compared against SGDM, RMSpop, Adam, AdamW, diffGrad and AdaBelief, implementing different deep architectures, in particular: (a) ResNet18, (b) ResNet50, (c) ResNet101 and their weight-standardised counterparts, i.e. (d) ResNet18ws, (e) ResNet50ws and (f) ResNet101ws, respectively.}
\label{fig:MINI_IMAGENET_new}
\end{figure*}

%%%%%%%%%%%%%%%%%%%%
\begin{table*}[!t]
\let\center\empty
\let\endcenter\relax
\centering
\caption{Results ($|\mu\pm\sigma|$) on the Mini-ImageNet data set in terms of Overall Accuracy and standard deviation. ResNet18, ResNet50, ResNet101 and their weight-standardised counterparts (ResNet18ws, ResNet50ws and ResNet101ws) have been considered.}
\resizebox{0.9\linewidth}{!}{

\input{tables/mini-Imagenet.tex}}
\label{table:compMINIimagenet}
\end{table*}

\begin{figure*}[!ht]
\centering
\includegraphics[width=\textwidth]{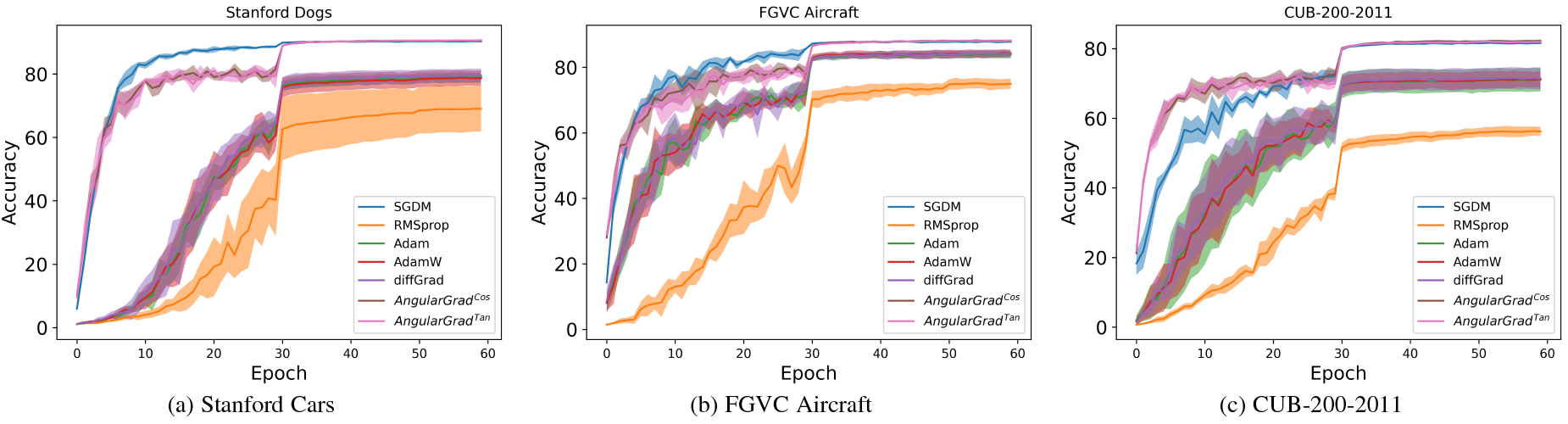}
\caption{Classification overall accuracy (OA) over the (a) Stanford Cars, (b) FGVC Aircraft and (c) CUB-200-2011 data sets.}
\label{fig:finegradACC}
\end{figure*}

\subsubsection{Classification results on Mini-ImageNet}
%%%%%%
    %The second experiment also tackles the classification problem with the aim of evaluating the OA of the proposed method over Mini-ImageNet dataset \cite{vinyals2016matching}. This challenging dataset is a subset of ImageNet dataset \cite{russakovsky2015imagenet} and includes 60000 $84\times84$ RGB images labelled into 100 different classes (with 600 images each class). Moreover, this dataset is divided into three subsets, one for training purposes (which comprises 64 classes), one for validation (comprising 16 classes), and another one for testing (composed by 20 classes). 
    %%
   % ResNet50 has been implemented, embedding SGDM, RMSprop, Adam, AdamW and DiffGrad, in addition to the proposed optimizer based on both $\cos$ and $\sin$ measurements.
    %%%%%%%%%%%%%%%%%%%%%%%%%%%%%%%%%%%%%
    %%%%%%%%%%%%%%%%%%%%%%%%%%%%%%%%%%%%%
%%%%%%
To further validate the \texttt{AngularGrad} optimizer, we retrain all the well-known deep architectures with different optimization methods on the Mini-ImageNet data set \cite{vinyals2016matching}. This data set is extracted from ImageNet set \cite{russakovsky2015imagenet},  it comprises 100 different categories or classes, where each one is composed of 600 images. Inspired by some previous works \cite{ravi2016optimization,iscen2019label,yong2020gradient}, 500 images have been randomly selected for training purposes, while the remaining 100 images are used for testing. As a result, we employ a training and testing sets of 50k and 10k images, respectively to study the performance of the proposed optimizer. 

Table~\ref{table:compMINIimagenet} reports the mean OAs from 5 runs (with random initialization of the network weights) for all the optimizers, evaluated on the considered data set. Indeed, the proposed \texttt{AngularGrad} and the other optimizers (SGDM, RMSprop, Adam, AdamW, diffGrad and AdaBelief) are evaluated with the same settings as those described in the first experiment. The learning rate scheduler divides the learning rate by 10 in the 80th epoch. Again, data augmentation is applied\footnote{
%%%%% FOOT NOTE STARTS
Images have been resized from $84\times84$ to $224\times224$ pixels and then flipped horizontally with probability of 50\%}.
%%% FOOTNOTE ENDS
It can be observed that \texttt{AngularGrad$^\tan$} and \texttt{AngularGrad$^\cos$} achieve the best accuracy results in all the deep models on which it has been run. Moreover, proposed algorithms generally exhibit a smaller standard deviation than the other optimization methods, where \texttt{AngularGrad$^\tan$} usually provides slightly better OA than \texttt{AngularGrad$^\cos$} as both achieve closer performance, which reduces the generalization gap between the two proposed optimization algorithms. The proposed methods outperform not only the widely-used SGD method but also the other considered adaptive methods. Graphically speaking, Fig.~\ref{fig:MINI_IMAGENET_new}(a)-(f) illustrates the faster and smoother convergence (with minimum variation achieved step-size) for the proposed algorithms as compared to the other considered methods over ResNet18 (a), ResNet50 (b), ResNet101 (c) and their weight-standardized counterparts (e)-(f).

\begin{figure*}[!t]
\centering
\includegraphics[width=\textwidth]{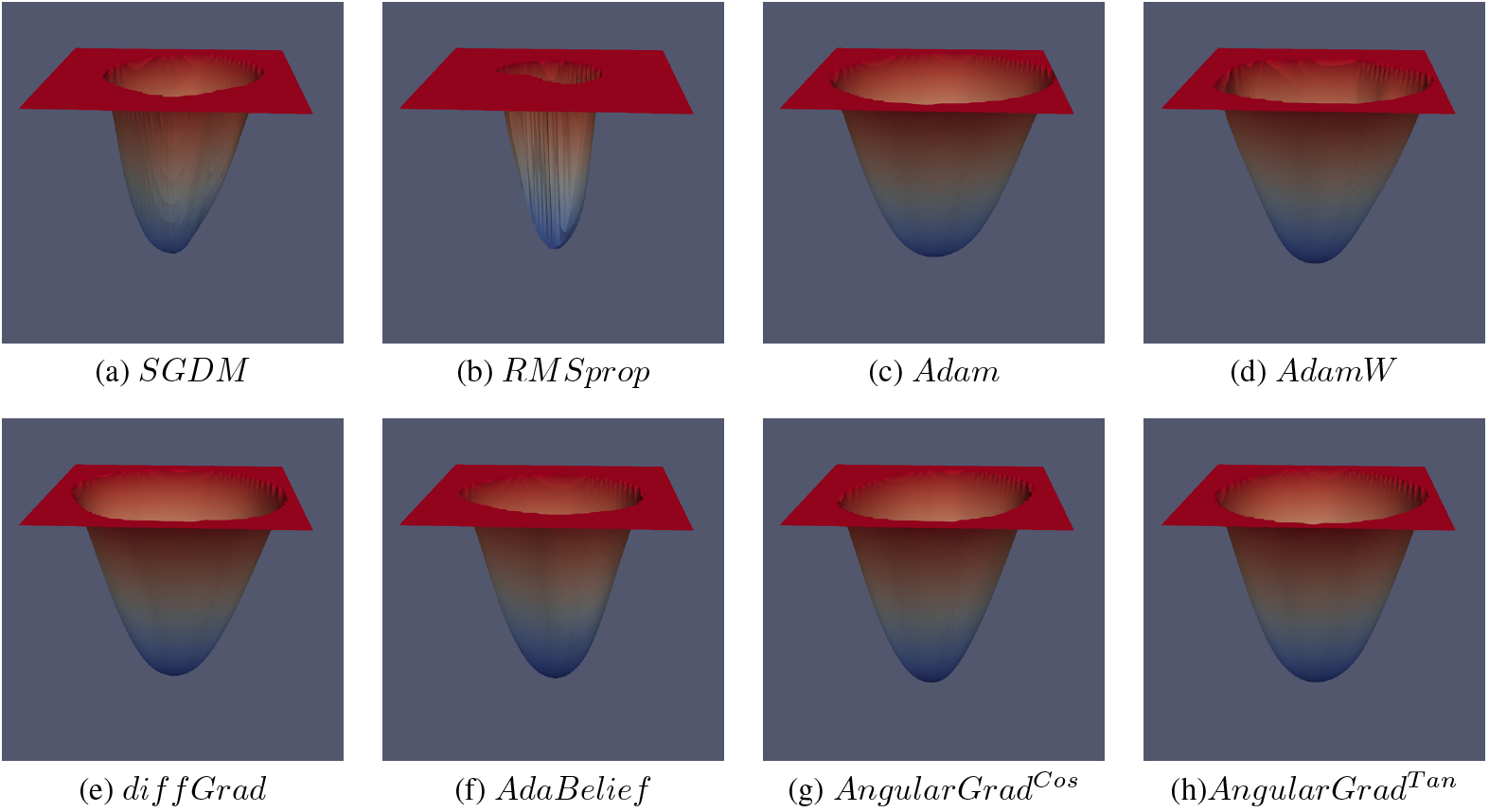}
\caption{3D graphical visualization of the Loss obtained on the Mini-Imagenet data set using different optimization algorithms: (a) SGDM, (b) RMSprop, (c) Adam, (d) AdamW, (e) diffGrad, (f) AdaBelief, and the proposed (g) AngularGrad$^{Cos}$  and (h) AngularGrad$^{Tan}$ considering ResNet50 as the deep network architecture.}
\label{fig:losslandscape}
\end{figure*}

\subsubsection{Classification Results on ImageNet}
%%%%%%%%%%%%%%%%%%%%
To further explore the proposed \texttt{AngularGrad} optimizer on classification tasks, this experiment delves into classification performance through ImageNet dataset \cite{krizhevsky2009learning,NIPS2012_c399862d}, using a ResNet-18 classifier. This classification benchmark comprises almost 15 million hand-tagged high-resolution images (concretely 14,197,122 scenes) belonging to 21,841 different categories. This dataset is more challenging to classify as it is more diverse and larger in scale compared to other classification marks. Following the ImageNet Large Scale Visual Recognition Challenge (ILSVRC) 2012 \cite{russakovsky2015imagenet}, a subset of ImageNet is considered to conduct the experimentation, i.e., 1.28 million variable-resolution labeled images of 1000 categories have been considered for training. Also, 50,000 labeled images have been used for validation and 100,000 unlabeled images have been considered for testing. Furthermore, images are down-sampled to a fixed resolution of $256\times256\times3$, and, then cropped to random patches of size $224\times224\times3$. % in 2010, as part of the Pascal Visual Object Challenge, an annual competition called the ImageNet Large- Scale Visual Recognition Challenge (ILSVRC) has been held. ILSVRC uses a subset of ImageNet with roughly 1000 images in each of 1000 categories. In all, there are roughly 1.2 million training images, 50,000 validation images, and 150,000 testing images.

\begin{table*}[!t]
\let\center\empty
\let\endcenter\relax
\centering
\caption{Top-1 accuracy of ResNet18 on ImageNet. $\dag$ is reported in \cite{chen2018closing}, $\ddag$ is reported in \cite{liu2019radam}, $\S$ is reported in \cite{zhuang2020adabelief}}
\resizebox{0.9\linewidth}{!}{

\input{tables/imagenet.tex}}
\label{table:compImagenet}
\end{table*}

By the same token, the proposed \texttt{AngularGrad} optimizer have been set with a learning rate of 0.005 and a weigh decay of 0.01, training with a batch size of 256 samples and 90 epochs. Moreover, the learning rate decay is done in epochs 50 and 70, dividing by 10. Regarding the other optimization methods, and with the aim of conducting a fair comparison, the best configuration of each of the optimized systems has been taken, following previous works to report the best result in the literature \cite{zhuang2020adabelief,chen2018closing,liu2019radam}. Consequently, configurations of AdaBelief and MSV AG have been provided by Zhuang \emph{et al.} in \cite{zhuang2020adabelief}, SGDM, AdaBound Yogi, and AdamW have been successfully reported by Chen \emph{et al.} in \cite{chen2018closing}, and RAdam has been studied by Liu \emph{et al.} in \cite{liu2019radam}. Finally, the configurations of Adam optimizer given by \cite{chen2018closing} and \cite{liu2019radam} are considered, too. 

Table \ref{table:compImagenet} provides the comparison between the proposed \texttt{AngularGrad} and the other considered optimization methods in terms of accuracy. It is noticeable that the proposed method obtains the best classification result on the ImageNet dataset (which is highlighted in bold), outperforming the SGDM method. Indeed, only \texttt{AngularGrad} (\textbf{70.25\%}), SGDM (70.23\%) and AdaBelief (which achieves the 70.08\% exploiting curvature information) are able to surpass the 70\% accuracy, where Adam and MSV AG optimizers provide the lowest accuracies (63.23\%-66.54\% and 65.99\%, respectively). Notwithstanding the accuracy results provided by Yogi (68.28\%), AdaBound (68.13\%) and AdamW (67.93\%), these methods are still hindered by a large generalization gap on the ImageNet dataset. In this context, and in contrast to the other optimization strategies, the trajectory followed by the gradient computed through the proposed \texttt{AngularGrad} method improves the performance of the ResNet-18 classifier on a dataset as complex as the ImageNet benchmark, not only closing the well-known generalization gap between SGDM and the other adaptive methods, but even outperforming SGDM.

\subsubsection{Fine-grained classification results}
%%%%%%   
    %The third experiment 
    %%%%%%%%%%%%%%%%%%%%%%%%%%%%%%%%%%%%%
    %%%%%%%%%%%%%%%%%%%%%%%%%%%%%%%%%%%%%
The fifth experiment is drawn to test the performance of the proposed optimization algorithm with pre-trained models for fine-grained classification. In this sense, SGDM is implemented with $\mu=0.1$ and the remaining algorithms with $\mu=1e-3$. Also, for each optimizer the feature extractor sets its learning rate to $\mu/10$, while the last layer keeps it to $\mu$. Furthermore, the learning rate scheduler divides the learning rate by 10 in the 30th and 50th epochs. Once more, data augmentation is applied\footnote{
%%%% foot note starts
Images have been resized to $512\times512$ pixels and then flipped horizontally with probability of 50\%. Also, the image contrast and brightness have been randomly changed. Finally, random cropping is applied to obtain $448\times448$ input images.}.%%%% foot note ends
Table~\ref{table:compFinedGrained} shows the OAs obtained over the test data considered for fine-grained image classification. It can be observed from the reported results that both the \texttt{AngularGrad$^\cos$}, and \texttt{AngularGrad$^\tan$} perform  significantly better than all the compared optimizers on these four data sets. However, the performance is slightly improved as compared to SDGM (81.47$\pm$0.23 vs. 81.21$\pm$0.54) on the CUB data set. Fig.~\ref{fig:finegradACC}(a)-(\textcolor{blue}{c}) illustrate the faster and smoother OA convergence (with minimum variation) achieved by the proposed optimizer as compared to the other tested methods methods over the Stanford Cars, Stanford Dogs, FGVC Aircraft and CUB-200-2011 data sets.

%%%%%%%%%%%%%%%%%%%%%
% fine-grained table
\begin{table}[!t]
\let\center\empty
\let\endcenter\relax
\centering
\caption{Results ($|\mu\pm\sigma|$) on Fine-Grained data sets with ResNet50 using different optimizers.}
\resizebox{0.95\linewidth}{!}{\input{tables/fined-gradient}}
\label{table:compFinedGrained}
\end{table}

%%%%%%%%%%%%%%%%%%%%%%%
\subsubsection{Analysis of Loss Landscape}
It is a difficult task to minimize a high-dimensional and non-convex loss function through the training process of DNNs \cite{blum1992training}, but sometimes it becomes easier in practical scenarios. Generally, SGD \cite{zhang2016understanding} is used to find optimal parameter solutions or saddle points where training loss is zero or near-zero, even when both the data and their associated labels are randomly shuffled before training. The training ability of any deep CNN directly or indirectly relies on various factors, including network architecture, activation function, network weight initialization,and most importantly, the choice of the optimizer for the defined network configuration. In order to visualize the effects using different optimizers on the network architecture, the ResNet50 model is trained using SGDM, RMSProp, Adam, AdamW, diffGrad, AdaBelief, and \texttt{AngularGrad$^\cos$}, and \texttt{AngularGrad$^\tan$} over the Mini-Imagenet data set. A visualization of the 3D loss landscape \cite{li2017visualizing} trajectory of the ResNet50 for different optimizers is shown in Fig.~\ref{fig:losslandscape}. It can be observed that the loss landscapes of SGDM, RMSprop, Adam and AdamW are not completely uniform, i.e., they are more steep in one side and less steep in another side. However, the loss landscapes of diffGrad, AdaBelief, \texttt{AngularGrad$^\cos$} and \texttt{AngularGrad$^\tan$} are more or less uniform in shape.
%%%%%%%%%%%%%%%%%%%%%%%

%%%%%%%%%%%%%%%%%%%%%%%
\section{Conclusion}
\label{concl}
In this paper, we propose a new \texttt{AngularGrad} optimizer that generates an angular coefficient to adaptively control the step size in the convergence process. It exploits gradient angular information between different iterations. To the best of our knowledge, this is the first attempt in the literature to use the angular information of the gradient for optimization purposes.
% existing optimizers did not utilize the direction/angle information of gradient vector efficiently. 
The proposed \texttt{AngularGrad} optimizer operates by considering the angle between two consecutive gradient vectors. The proposed optimizer generates a more accurate step size and achieves faster and smoother convergence. A good generalization shows training stability in complex settings for both image and fine-grained classification tasks. Comprehensive experiments have been conducted in different computer vision tasks with different optimizers and well-known network architectures. The obtained results demonstrate that the newly proposed \texttt{AngularGrad} optimizer improves training efficiency and performance. Further experiments with additional data sets should be conducted to fully validate the aforementioned concluding remarks.

% {\appendix[Proof of the Zonklar Equations]
% Use $\backslash${\tt{appendix}} if you have a single appendix:
% Do not use $\backslash${\tt{section}} anymore after $\backslash${\tt{appendix}}, only $\backslash${\tt{section*}}.
% If you have multiple appendixes use $\backslash${\tt{appendices}} then use $\backslash${\tt{section}} to start each appendix.
% You must declare a $\backslash${\tt{section}} before using any $\backslash${\tt{subsection}} or using $\backslash${\tt{label}} ($\backslash${\tt{appendices}} by itself
%  starts a section numbered zero.)}


\bibliographystyle{IEEEtran}
\bibliography{main}

% \newpage

 

% \end{IEEEbiography}



% \vfill

\end{document}

%% file: tables/cifar10.tex
% \begin{tabular}{|c|ccccccc|}
% \hline
% Optimizer      & VGG11 & ResNet18 & ResNet34 & ResNet50 & ResNet101 & ResNeXt29 & DenseNet121 \\
% \hline
\begin{tabular}{|c|cccccccc|}
\hline
Optimizer      & VGG16 & ResNet18 & ResNet34 & ResNet50 & ResNet101 & ResNeXt29 & DLA & DenseNet121 \\
\hline

$SGDM$              & 92.45 (0.09) & 93.23 (0.14) & 93.42 (0.13) & 93.40 (0.19) & 92.92 (0.05) & 93.53 (0.11) & 93.39 (0.14) & 93.73 (0.09)  \\
$RMSprop$           & 92.35 (0.18) & 93.58 (0.20) & 93.85 (0.08) & 93.51 (0.18) & 93.77 (0.11) & 93.25 (0.12) & 93.09 (0.14) & 93.74 (0.11)  \\
$Adam$              & 92.74 (0.07) & 93.65 (0.21) & \textcolor{blue}{94.09} (0.18) & 93.93 (0.10) & 94.26 (0.07) & 93.46 (0.12) & 93.59 (0.14) & 94.15 (0.08)  \\
$AdamW$             & 92.70 (0.12) & \textcolor{blue}{93.74} (0.08) & 94.04 (0.15) & 94.12 (0.12) & 94.16 (0.18) & 93.58 (0.13) & 93.71 (0.20) & 94.20 (0.15)  \\
$diffGrad$          & 92.61 (0.05) & 93.70 (0.06) & 93.86 (0.23) & 93.88 (0.15) & \textcolor{blue}{94.28} (0.23) & 93.50 (0.19) & 93.47 (0.26) & 94.10 (0.13)  \\
$AdaBelief$         & 92.65 (0.11) & 93.51 (0.13) & 93.83 (0.16) & 93.83 (0.15) & 94.12 (0.14) & 93.83 (0.16) & 93.71 (0.10) & 94.15 (0.21)  \\
\hdashline
$AngularGrad^{Cos}$ & \textbf{92.87} (0.27) & \textbf{93.83} (0.12) & \textbf{94.14} (0.10) & \textbf{94.26} (0.15) & \textbf{94.35} (0.16) & \textcolor{blue}{93.89} (0.09) & \textcolor{blue}{93.74} (0.19) & \textcolor{blue}{94.42} (0.13)  \\
$AngularGrad^{Tan}$ & \textcolor{blue}{92.76} (0.14) & \textcolor{blue}{93.74} (0.12) & 94.06 (0.07) & \textcolor{blue}{94.22} (0.16) & 94.22 (0.16) & \textbf{93.98} (0.16) & \textbf{93.83} (0.16) & \textbf{94.43} (0.10)  \\

\hline
\end{tabular}

%% file: tables/cifar100.tex
% \begin{tabular}{|c|ccccccc|}
% \hline
% Optimizer      & VGG11 & ResNet18 & ResNet34 & ResNet50 & ResNet101 & ResNeXt29 & DenseNet121 \\
% \hline
\begin{tabular}{|c|cccccccc|}
\hline
Optimizer      & VGG16 & ResNet18 & ResNet34 & ResNet50 & ResNet101 & ResNeXt29 & DLA & DenseNet121 \\
\hline

$SGDM$              & \textbf{69.80} (0.22) & \textcolor{blue}{73.15} (0.29) & 73.20 (0.23) & 73.79 (0.29) & 72.70 (0.51) & \textcolor{blue}{75.88} (0.25) & \textbf{73.96} (0.25) & 74.90 (0.14)  \\
$RMSprop$           & 66.05 (0.63) & 70.02 (0.36) & 70.25 (0.58) & 70.97 (0.69) & 72.26 (0.85) & 71.58 (0.74) & 69.38 (1.16) & 72.33 (0.35)  \\
$Adam$              & 67.91 (0.19) & 71.73 (0.22) & 72.20 (0.21) & 73.83 (0.34) & 74.89 (0.29) & 73.16 (0.53) & 71.40 (0.47) & 74.45 (0.25)  \\
$AdamW$             & 67.78 (0.36) & 71.89 (0.27) & 72.26 (0.39) & 73.72 (0.47) & 75.45 (0.43) & 73.54 (0.34) & 71.75 (0.40) & 74.53 (0.32)  \\
$DiffGrad$          & 67.68 (0.24) & 71.56 (0.13) & 72.17 (0.27) & 73.48 (0.44) & 74.90 (0.72) & 73.22 (0.55) & 71.73 (0.09) & 74.42 (0.37)  \\
$AdaBelief$         & \textcolor{blue}{69.43} (0.28) & 73.02 (0.21) & 73.12 (0.29) & 75.17 (0.27) & \textbf{75.72} (0.17) & \textbf{75.94} (0.29) & \textcolor{blue}{73.94} (0.30) & 75.51 (0.21)  \\
\hdashline
$AngularGrad^{Cos}$ & 69.12 (0.17) & 72.90 (0.20) & \textcolor{blue}{73.42} (0.18) & \textbf{75.67} (0.29) & \textcolor{blue}{75.66} (0.21) & 75.41 (0.49) & 73.26 (0.16) & \textbf{75.55} (0.21)  \\
$AngularGrad^{Tan}$ & 69.05 (0.20) & \textbf{73.23} (0.27) & \textbf{73.62} (0.26) & \textcolor{blue}{75.52} (0.07) & 75.54 (0.23) & 75.56 (0.19) & 73.34 (0.17) & \textcolor{blue}{75.52} (0.10)  \\

\hline
\end{tabular}

%% file: tables/mini-Imagenet.tex
\begin{tabular}{|c|cccccc||cc|}
\hline
Model & $SGDM$ & $RMSprop$ & $Adam$ & $AdamW$ & $diffGrad$ & $AdaBelief$ & $AngularGrad^{Cos}$ & $AngularGrad^{Tan}$\\
\hline

ResNet18  & 65.11 (0.36)  & 63.89 (0.33) & 64.93 (0.24) & 65.21 (0.36) & 64.93 (0.30) & 65.70 (0.10) & \textbf{65.72} (0.23) & 65.54 (0.10)  \\
ResNet50  & 65.33 (0.43)  & 65.52 (0.54) & 66.91 (0.62) & 67.34 (0.33) & 66.93 (0.40) & 66.61 (0.19) & 68.15 (0.10) & \textbf{68.40} (0.19)  \\
ResNet101 & 64.37 (0.38) & 66.5 (0.44) & 68.42 (0.37) & 68.86 (0.43) & 68.68 (0.38) & 67.97 (0.47) & \textbf{69.29} (0.22) & 69.28 (0.22)  \\
\hline
ResNet18-WS  & 65.46 (0.10) & 65.28 (0.31) & 66.04 (0.27) & 66.63 (0.37) & 66.22 (0.07) & 66.40 (0.20) & \textbf{66.80} (0.21) & 66.64 (0.24)  \\
ResNet50-WS  & 65.66 (0.27) & 67.65 (0.42) & 68.98 (0.18) & 69.21 (0.33) & 68.89 (0.39) & 67.95 (0.44) & \textbf{69.45} (0.17) & 69.41 (0.16)  \\
ResNet101-WS & 65.47 (0.31) & 69.07 (0.31) & 69.78 (0.21) & 70.16 (0.25) & 69.82 (0.25) & 68.82 (0.30) & 70.18 (0.33) & \textbf{70.31} (0.23)  \\
% DenseNet121 & 48.54 (24.27) & 66.53 (0.27) & 67.21 (0.22) & 67.84 (0.25) & 67.48 (0.36) & 64.44 (0.25) & 66.26 (0.17) & 66.21 (0.3)  \\
% DenseNet161 & 61.52 (0.2) & 67.77 (0.56) & 68.54 (0.39) & 69.09 (0.28) & 68.95 (0.41) & 65.24 (0.24) & 68.89 (0.21) & 68.84 (0.3)  \\
\hline
\end{tabular}

%% file: tables/imagenet.tex
\begin{tabular}{|c|cccccccc|}
\hline
$AngularGrad^{Tan}$ & $AdaBelief$ & $SGDM$ & $AdaBound$ & $Yogi$ & $Adam$ & $MSVAG$ & $RAdam$ & $AdamW$ \\
\hline
\textbf{70.25}  & 70.08$^\S$ & 70.23$^\dag$ & 68.13$^\dag$ & 68.23$^\dag$ & 63.79$^\dag$ (66.54$^\ddag$) & 65.99$^\S$ & 67.62$^\ddag$ & 67.93$^\dag$ \\
\hline
\end{tabular}

%% file: tables/fined-gradient.tex
% \begin{tabular}{|c|cccc|}
% \hline
% Optimizer      & Cars & Dogs & Aircraft & CUB \\
% \hline
% % $SGDM$              & 89.62 (0.11) & 78.35 (0.18) & 86.84 (0.41) & 81.21 (0.54)  \\
% % $RMSprop$           & 68.77 (4.92) & 61.84 (1.22) & 73.81 (0.96) & 55.04 (1.10)  \\
% % $Adam$              & 79.70 (1.25) & 76.12 (0.22) & 84.07 (0.81) & 71.11 (3.69)  \\
% % $AdamW$             & 77.79 (2.35) & 76.30 (0.31) & 84.36 (0.81) & 71.13 (2.80)  \\
% % $diffGrad$          & 78.32 (2.00) & 76.44 (0.50) & 84.21 (0.62) & 71.44 (3.03)  \\
% % $AngularGrad^{Cos}$ & 90.23 (0.31) & \textbf{79.84} (0.25) & 88.09 (0.32) & \textbf{81.47} (0.23)  \\
% % $AngularGrad^{Tan}$ & \textbf{90.30} (0.18) & 79.68 (0.34) & \textbf{88.18} (0.44) & 81.28 (0.54)  \\

% $SGDM$              & 90.43(0.11) & \textbf{86.29}(0.17) & 88.10(0.25) & 81.81(0.27)  \\
% $RMSprop$           & 69.19(7.21) & 63.37(1.40) & 75.32(1.59) & 56.51(1.51)  \\
% $Adam$              & 79.05(1.57) & 76.65(0.50) & 84.09(1.12) & 71.24(3.34)  \\
% $Adamw$             & 78.88(2.46) & 76.78(0.29) & 84.38(1.16) & 71.38(2.34)  \\
% $diffGrad$          & 79.72(2.03) & 76.49(0.39) & 84.32(1.17) & 71.74(3.11)  \\
% $AngularGrad^{Cos}$ & \textbf{90.71}(0.20) & 79.83(0.21) & \textbf{88.30}(0.26) & \textbf{82.45}(0.34)  \\
% $AngularGrad^{Tan}$ & 90.70(0.19) & 80.11(0.29) & 88.23(0.25) & 82.21(0.22)  \\

% \hline
% % $AdaBelief$ & 91.2 (0.0) & 83.01 (0.0) & 88.28 (0.0) & 84.65 (0.0)  \\
% % \hline
% \end{tabular}

\begin{tabular}{|c|ccc|}
\hline
Optimizer      & Cars & Aircraft & CUB \\
\hline
$SGDM$              & 90.43(0.11) & 88.10(0.25) & 81.81(0.27)  \\
$RMSprop$           & 69.19(7.21) & 75.32(1.59) & 56.51(1.51)  \\
$Adam$              & 79.05(1.57) & 84.09(1.12) & 71.24(3.34)  \\
$Adamw$             & 78.88(2.46) & 84.38(1.16) & 71.38(2.34)  \\
$diffGrad$          & 79.72(2.03) & 84.32(1.17) & 71.74(3.11)  \\
\hdashline
$AngularGrad^{Cos}$ & \textbf{90.71}(0.20) & \textbf{88.30}(0.26) & \textbf{82.45}(0.34)  \\
$AngularGrad^{Tan}$ & 90.70(0.19) & 88.23(0.25) & 82.21(0.22)  \\

\hline
% $AdaBelief$ & 91.2 (0.0) & 83.01 (0.0) & 88.28 (0.0) & 84.65 (0.0)  \\
% \hline
\end{tabular}

%% file: main.bbl
\begin{thebibliography}{10}
\providecommand{\url}[1]{#1}
\csname url@samestyle\endcsname
\providecommand{\newblock}{\relax}
\providecommand{\bibinfo}[2]{#2}
\providecommand{\BIBentrySTDinterwordspacing}{\spaceskip=0pt\relax}
\providecommand{\BIBentryALTinterwordstretchfactor}{4}
\providecommand{\BIBentryALTinterwordspacing}{\spaceskip=\fontdimen2\font plus
\BIBentryALTinterwordstretchfactor\fontdimen3\font minus \fontdimen4\font\relax}
\providecommand{\BIBforeignlanguage}[2]{{%
\expandafter\ifx\csname l@#1\endcsname\relax
\typeout{** WARNING: IEEEtran.bst: No hyphenation pattern has been}%
\typeout{** loaded for the language `#1'. Using the pattern for}%
\typeout{** the default language instead.}%
\else
\language=\csname l@#1\endcsname
\fi
#2}}
\providecommand{\BIBdecl}{\relax}
\BIBdecl

\bibitem{moore1998cramming}
G.~E. Moore, ``Cramming more components onto integrated circuits,'' \emph{Proceedings of the IEEE}, vol.~86, no.~1, pp. 82--85, 1998.

\bibitem{chen2014data}
C.~P. Chen and C.-Y. Zhang, ``Data-intensive applications, challenges, techniques and technologies: A survey on big data,'' \emph{Information sciences}, vol. 275, pp. 314--347, 2014.

\bibitem{lecun2015deep}
Y.~LeCun, Y.~Bengio, and G.~Hinton, ``Deep learning,'' \emph{nature}, vol. 521, no. 7553, pp. 436--444, 2015.

\bibitem{goodfellow2016deep}
I.~Goodfellow, Y.~Bengio, and A.~Courville, \emph{Deep Learning}.\hskip 1em plus 0.5em minus 0.4em\relax MIT Press, 2016, \url{http://www.deeplearningbook.org}.

\bibitem{bishop2006pattern}
C.~M. Bishop, \emph{Pattern recognition and machine learning}.\hskip 1em plus 0.5em minus 0.4em\relax springer, 2006.

\bibitem{gopisetty2008evolution}
S.~Gopisetty, S.~Agarwala, E.~Butler, D.~Jadav, S.~Jaquet, M.~Korupolu, R.~Routray, P.~Sarkar, A.~Singh, M.~Sivan-Zimet \emph{et~al.}, ``Evolution of storage management: Transforming raw data into information,'' \emph{IBM Journal of Research and Development}, vol.~52, no. 4.5, pp. 341--352, 2008.

\bibitem{schmidhuber2015deep}
J.~Schmidhuber, ``Deep learning in neural networks: An overview,'' \emph{Neural networks}, vol.~61, pp. 85--117, 2015.

\bibitem{sabour2017dynamic}
S.~Sabour, N.~Frosst, and G.~E. Hinton, ``Dynamic routing between capsules,'' in \emph{Advances in neural information processing systems}, 2017, pp. 3856--3866.

\bibitem{cirecsan2012deep}
D.~C. Cire{\c{s}}an, U.~Meier, L.~M. Gambardella, and J.~Schmidhuber, ``Deep big multilayer perceptrons for digit recognition,'' in \emph{Neural networks: tricks of the trade}.\hskip 1em plus 0.5em minus 0.4em\relax Springer, 2012, pp. 581--598.

\bibitem{zabalza2016novel}
J.~Zabalza, J.~Ren, J.~Zheng, H.~Zhao, C.~Qing, Z.~Yang, P.~Du, and S.~Marshall, ``Novel segmented stacked autoencoder for effective dimensionality reduction and feature extraction in hyperspectral imaging,'' \emph{Neurocomputing}, vol. 185, pp. 1--10, 2016.

\bibitem{hinton2009deep}
G.~E. Hinton, ``Deep belief networks,'' \emph{Scholarpedia}, vol.~4, no.~5, p. 5947, 2009.

\bibitem{NIPS2012_c399862d}
A.~Krizhevsky, I.~Sutskever, and G.~E. Hinton, ``Imagenet classification with deep convolutional neural networks,'' in \emph{Advances in Neural Information Processing Systems}, F.~Pereira, C.~J.~C. Burges, L.~Bottou, and K.~Q. Weinberger, Eds., vol.~25.\hskip 1em plus 0.5em minus 0.4em\relax Curran Associates, Inc., 2012, pp. 1097--1105.

\bibitem{pascanu2013difficulty}
R.~Pascanu, T.~Mikolov, and Y.~Bengio, ``On the difficulty of training recurrent neural networks,'' in \emph{International conference on machine learning}, 2013, pp. 1310--1318.

\bibitem{he2016deep}
K.~He, X.~Zhang, S.~Ren, and J.~Sun, ``Deep residual learning for image recognition,'' in \emph{Proceedings of the IEEE conference on computer vision and pattern recognition}, 2016, pp. 770--778.

\bibitem{huang2017densely}
G.~Huang, Z.~Liu, L.~Van Der~Maaten, and K.~Q. Weinberger, ``Densely connected convolutional networks,'' in \emph{Proceedings of the IEEE conference on computer vision and pattern recognition}, 2017, pp. 4700--4708.

\bibitem{szegedy2016rethinking}
C.~Szegedy, V.~Vanhoucke, S.~Ioffe, J.~Shlens, and Z.~Wojna, ``Rethinking the inception architecture for computer vision,'' in \emph{Proceedings of the IEEE conference on computer vision and pattern recognition}, 2016, pp. 2818--2826.

\bibitem{wang2017residual}
F.~Wang, M.~Jiang, C.~Qian, S.~Yang, C.~Li, H.~Zhang, X.~Wang, and X.~Tang, ``Residual attention network for image classification,'' in \emph{Proceedings of the IEEE conference on computer vision and pattern recognition}, 2017, pp. 3156--3164.

\bibitem{chen2017dual}
Y.~Chen, J.~Li, H.~Xiao, X.~Jin, S.~Yan, and J.~Feng, ``Dual path networks,'' in \emph{Advances in neural information processing systems}, 2017, pp. 4467--4475.

\bibitem{goodfellow2014generative}
I.~J. Goodfellow, J.~Pouget-Abadie, M.~Mirza, B.~Xu, D.~Warde-Farley, S.~Ozair, A.~Courville, and Y.~Bengio, ``Generative adversarial networks,'' \emph{Advances in neural information processing systems}, vol.~3, no.~06, 2014.

\bibitem{zhang2019deeper}
Z.~Zhang and H.~Peng, ``Deeper and wider siamese networks for real-time visual tracking,'' in \emph{Proceedings of the IEEE Conference on Computer Vision and Pattern Recognition}, 2019, pp. 4591--4600.

\bibitem{mirzaei2020deep}
A.~Mirzaei, V.~Pourahmadi, M.~Soltani, and H.~Sheikhzadeh, ``Deep feature selection using a teacher-student network,'' \emph{Neurocomputing}, vol. 383, pp. 396--408, 2020.

\bibitem{yu2010deep}
D.~Yu and L.~Deng, ``Deep learning and its applications to signal and information processing [exploratory dsp],'' \emph{IEEE Signal Processing Magazine}, vol.~28, no.~1, pp. 145--154, 2010.

\bibitem{young2018recent}
T.~Young, D.~Hazarika, S.~Poria, and E.~Cambria, ``Recent trends in deep learning based natural language processing,'' \emph{IEEE Computational intelligence magazine}, vol.~13, no.~3, pp. 55--75, 2018.

\bibitem{szegedy2013deep}
C.~Szegedy, A.~Toshev, and D.~Erhan, ``Deep neural networks for object detection,'' in \emph{Advances in neural information processing systems}, 2013, pp. 2553--2561.

\bibitem{wang2013learning}
N.~Wang and D.-Y. Yeung, ``Learning a deep compact image representation for visual tracking,'' in \emph{Advances in neural information processing systems}, 2013, pp. 809--817.

\bibitem{pierson2017deep}
H.~A. Pierson and M.~S. Gashler, ``Deep learning in robotics: a review of recent research,'' \emph{Advanced Robotics}, vol.~31, no.~16, pp. 821--835, 2017.

\bibitem{wang2018interactive}
G.~Wang, W.~Li, M.~A. Zuluaga, R.~Pratt, P.~A. Patel, M.~Aertsen, T.~Doel, A.~L. David, J.~Deprest, S.~Ourselin \emph{et~al.}, ``Interactive medical image segmentation using deep learning with image-specific fine tuning,'' \emph{IEEE transactions on medical imaging}, vol.~37, no.~7, pp. 1562--1573, 2018.

\bibitem{paoletti2019deep}
M.~Paoletti, J.~Haut, J.~Plaza, and A.~Plaza, ``Deep learning classifiers for hyperspectral imaging: A review,'' \emph{ISPRS Journal of Photogrammetry and Remote Sensing}, vol. 158, pp. 279--317, 2019.

\bibitem{simonyan2014very}
K.~Simonyan and A.~Zisserman, ``Very deep convolutional networks for large-scale image recognition,'' \emph{arXiv preprint arXiv:1409.1556}, 2014.

\bibitem{xie2017aggregated}
S.~Xie, R.~Girshick, P.~Doll{\'a}r, Z.~Tu, and K.~He, ``Aggregated residual transformations for deep neural networks,'' in \emph{Proceedings of the IEEE conference on computer vision and pattern recognition}, 2017, pp. 1492--1500.

\bibitem{csaji2001approximation}
B.~C. Cs{\'a}ji \emph{et~al.}, ``Approximation with artificial neural networks,'' \emph{Faculty of Sciences, Etvs Lornd University, Hungary}, vol.~24, no.~48, p.~7, 2001.

\bibitem{zhou2020universality}
D.-X. Zhou, ``Universality of deep convolutional neural networks,'' \emph{Applied and computational harmonic analysis}, vol.~48, no.~2, pp. 787--794, 2020.

\bibitem{yong2020gradient}
H.~Yong, J.~Huang, X.~Hua, and L.~Zhang, ``Gradient centralization: A new optimization technique for deep neural networks,'' \emph{arXiv preprint arXiv:2004.01461}, 2020.

\bibitem{bottou1991stochastic}
L.~Bottou, ``Stochastic gradient learning in neural networks,'' \emph{Proceedings of Neuro-N{\i}mes}, vol.~91, no.~8, p.~12, 1991.

\bibitem{qian1999momentum}
N.~Qian, ``On the momentum term in gradient descent learning algorithms,'' \emph{Neural networks}, vol.~12, no.~1, pp. 145--151, 1999.

\bibitem{hinton2012neural}
G.~Hinton, N.~Srivastava, and K.~Swersky, ``Neural networks for machine learning lecture 6a overview of mini-batch gradient descent,'' \emph{Cited on}, vol.~14, no.~8, 2012.

\bibitem{duchi2011adaptive}
J.~Duchi, E.~Hazan, and Y.~Singer, ``Adaptive subgradient methods for online learning and stochastic optimization.'' \emph{Journal of machine learning research}, vol.~12, no.~7, 2011.

\bibitem{kingma2014adam}
D.~P. Kingma and J.~Ba, ``Adam: A method for stochastic optimization,'' \emph{arXiv preprint arXiv:1412.6980}, 2014.

\bibitem{loshchilov2017decoupled}
I.~Loshchilov and F.~Hutter, ``Decoupled weight decay regularization,'' \emph{arXiv preprint arXiv:1711.05101}, 2017.

\bibitem{dubey2019diffgrad}
S.~Dubey, S.~Chakraborty, S.~Roy, S.~Mukherjee, S.~Singh, and B.~Chaudhuri, ``diffgrad: An optimization method for convolutional neural networks.'' \emph{IEEE transactions on neural networks and learning systems}, vol.~31, no.~11, pp. 4500--4511, 2020.

\bibitem{baydin2017online}
A.~G. Baydin, R.~Cornish, D.~M. Rubio, M.~Schmidt, and F.~Wood, ``Online learning rate adaptation with hypergradient descent,'' \emph{arXiv preprint arXiv:1703.04782}, 2017.

\bibitem{li2020rethinking}
H.~Li, P.~Chaudhari, H.~Yang, M.~Lam, A.~Ravichandran, R.~Bhotika, and S.~Soatto, ``Rethinking the hyperparameters for fine-tuning,'' \emph{arXiv preprint arXiv:2002.11770}, 2020.

\bibitem{dauphin2019metainit}
Y.~N. Dauphin and S.~Schoenholz, ``Metainit: Initializing learning by learning to initialize,'' \emph{Advances in Neural Information Processing Systems}, vol.~32, 2019.

\bibitem{sutskever2013importance}
I.~Sutskever, J.~Martens, G.~Dahl, and G.~Hinton, ``On the importance of initialization and momentum in deep learning,'' in \emph{International conference on machine learning}.\hskip 1em plus 0.5em minus 0.4em\relax PMLR, 2013, pp. 1139--1147.

\bibitem{smith2021origin}
S.~L. Smith, B.~Dherin, D.~G. Barrett, and S.~De, ``On the origin of implicit regularization in stochastic gradient descent,'' \emph{arXiv preprint arXiv:2101.12176}, 2021.

\bibitem{liu2019radam}
L.~Liu, H.~Jiang, P.~He, W.~Chen, X.~Liu, J.~Gao, and J.~Han, ``On the variance of the adaptive learning rate and beyond,'' in \emph{Proceedings of the Eighth International Conference on Learning Representations (ICLR 2020)}, April 2020, p.~.

\bibitem{zhuang2020adabelief}
J.~Zhuang, T.~Tang, Y.~Ding, S.~Tatikonda, N.~Dvornek, X.~Papademetris, and J.~Duncan, ``Adabelief optimizer: Adapting stepsizes by the belief in observed gradients,'' \emph{Conference on Neural Information Processing Systems}, 2020.

\bibitem{krizhevsky2009learning}
A.~Krizhevsky and G.~Hinton, ``Learning multiple layers of features from tiny images,'' \emph{Department of Computer Science, University of Toronto}, 2009.

\bibitem{vinyals2016matching}
O.~Vinyals, C.~Blundell, T.~Lillicrap, D.~Wierstra \emph{et~al.}, ``Matching networks for one shot learning,'' in \emph{Advances in neural information processing systems}, 2016, pp. 3630--3638.

\bibitem{qiao2019micro}
S.~Qiao, H.~Wang, C.~Liu, W.~Shen, and A.~Yuille, ``Micro-batch training with batch-channel normalization and weight standardization,'' \emph{arXiv preprint arXiv:1903.10520}, 2019.

\bibitem{chen2018closing}
J.~Chen and Q.~Gu, ``Closing the generalization gap of adaptive gradient methods in training deep neural networks,'' \emph{arXiv preprint arXiv:1806.06763}, 2018.

\bibitem{krause20133d}
J.~Krause, M.~Stark, J.~Deng, and L.~Fei-Fei, ``3d object representations for fine-grained categorization,'' in \emph{Proceedings of the IEEE international conference on computer vision workshops}, 2013, pp. 554--561.

\bibitem{khosla2011novel}
A.~Khosla, N.~Jayadevaprakash, B.~Yao, and F.-F. Li, ``Novel dataset for fgvc: Stanford dogs,'' in \emph{San Diego: CVPR Workshop on FGVC}, vol.~1, no.~2, 2011, p.~.

\bibitem{maji2013fine}
S.~Maji, E.~Rahtu, J.~Kannala, M.~Blaschko, and A.~Vedaldi, ``Fine-grained visual classification of aircraft,'' \emph{arXiv preprint arXiv:1306.5151}, 2013.

\bibitem{wah2011caltech}
C.~Wah, S.~Branson, P.~Welinder, P.~Perona, and S.~Belongie, ``The caltech-ucsd birds-200-2011 dataset,'' California Institute of Technology, Tech. Rep. CNS-TR-2011-001, 2011.

\bibitem{rosenbrock1960automatic}
H.~Rosenbrock, ``An automatic method for finding the greatest or least value of a function,'' \emph{The Computer Journal}, vol.~3, no.~3, pp. 175--184, 1960.

\bibitem{yu2018deep}
F.~Yu, D.~Wang, E.~Shelhamer, and T.~Darrell, ``Deep layer aggregation,'' in \emph{Proceedings of the IEEE conference on computer vision and pattern recognition}, 2018, pp. 2403--2412.

\bibitem{russakovsky2015imagenet}
O.~Russakovsky, J.~Deng, H.~Su, J.~Krause, S.~Satheesh, S.~Ma, Z.~Huang, A.~Karpathy, A.~Khosla, M.~Bernstein \emph{et~al.}, ``Imagenet large scale visual recognition challenge,'' \emph{International journal of computer vision}, vol. 115, no.~3, pp. 211--252, 2015.

\bibitem{ravi2016optimization}
S.~Ravi and H.~Larochelle, ``Optimization as a model for few-shot learning,'' in \emph{International conference on learning representations}, 2017, p.~.

\bibitem{iscen2019label}
A.~Iscen, G.~Tolias, Y.~Avrithis, and O.~Chum, ``Label propagation for deep semi-supervised learning,'' in \emph{Proceedings of the IEEE/CVF Conference on Computer Vision and Pattern Recognition}, 2019, pp. 5070--5079.

\bibitem{blum1992training}
A.~L. Blum and R.~L. Rivest, ``Training a 3-node neural network is np-complete,'' \emph{Neural Networks}, vol.~5, no.~1, pp. 117--127, 1992.

\bibitem{zhang2016understanding}
C.~Zhang, S.~Bengio, M.~Hardt, B.~Recht, and O.~Vinyals, ``Understanding deep learning requires rethinking generalization,'' \emph{arXiv preprint arXiv:1611.03530}, 2016.

\bibitem{li2017visualizing}
H.~Li, Z.~Xu, G.~Taylor, C.~Studer, and T.~Goldstein, ``Visualizing the loss landscape of neural nets,'' \emph{arXiv preprint arXiv:1712.09913}, 2017.

\end{thebibliography}
